\documentclass[journal]{IEEEtran}
\IEEEoverridecommandlockouts
\usepackage{graphicx}

\usepackage{amsfonts,amssymb}
\usepackage[ruled]{algorithm2e}
\usepackage{colortbl} 
\usepackage{xcolor}
\usepackage{subfigure}
\usepackage{amsmath}
\usepackage{mathrsfs}
\usepackage{bm}
\usepackage{epsfig}
\usepackage{rotating, multirow}
\usepackage{epstopdf}
\usepackage{float}
\usepackage{diagbox}
\usepackage{balance}
\usepackage{arydshln}
\usepackage{booktabs}

\newcommand{\tabincell}[2]{\begin{tabular}{@{}#1@{}}#2\end{tabular}}

\AtBeginDocument{%
  \providecommand\BibTeX{{%
    \normalfont B\kern-0.5em{\scshape i\kern-0.25em b}\kern-0.8em\TeX}}}

\ifCLASSINFOpdf

\else

\fi

\begin{document}
\title{Hierarchical Refinement of Universal Multimodal Attacks on Vision-Language Models}

\author{Peng-Fei Zhang, Guangdong Bai and Zi Huang, \textit{IEEE Fellow}
\thanks{This work is partially supported by Australian Research Council Discovery Project (DP230101196, CE200100025).}
\thanks{Peng-Fei Zhang, Zi Huang are with the School of Electrical Engineering and Computer Science, the University of Queensland, email: mima.zpf@gmail.com and huang@itee.uq.edu.au.}
\thanks{Guangdong Bai is with the Department of Computer Science, City University of Hong Kong, email: g.bai@cityu.edu.hk.}
}
\markboth{Journal of \LaTeX\ Class Files,~Vol.~14, No.~8, August~2015}%
{Shell \MakeLowercase{\textit{et al.}}: Bare Demo of IEEEtran.cls for IEEE Journals}

\maketitle

\begin{abstract}
Adversarial attacks, which apply imperceptible perturbations to mislead model predictions, are essential for evaluating the robustness of Vision–Language Pre-trained (VLP) models. However, existing adversarial attacks for VLP models are mostly sample-specific, resulting in substantial computational overhead when scaled to large datasets or new scenarios.

To overcome this limitation, we propose Hierarchical Refinement Attack (HRA), a multimodal universal attack framework for VLP models. For the image modality, we refine the optimization path by leveraging a temporal hierarchy of historical and estimated future gradients to avoid local minima and stabilize universal perturbation learning. For the text modality, it hierarchically models textual importance by considering both intra- and inter-sentence contributions to identify globally influential words, which are then used as universal text perturbations. Extensive experiments across various downstream tasks, VLP models, and datasets, demonstrate the superior transferability of the proposed universal multimodal attacks.
\end{abstract}


\IEEEpeerreviewmaketitle

\section{Introduction}
Vision-Language Pre-trained (VLP) models play a crucial role in bridging the gap between images and texts, enabling multimodal understanding, generation, and response \cite{radford2021learning,li2021align,yang2022vision}. Due to their promising performance and efficiency, VLP models have been widely applied to various vision-language tasks, ranging from multimodal retrieval \cite{liu2021image} to image captioning \cite{zhou2020unified}. As the deployment of VLP models grows, evaluating their robustness has become essential, particularly in high-stakes applications where reliability is critical. Uncovering model vulnerabilities would drive advancements in the development of more resilient and reliable neural networks. Adversarial attacks are a key strategy for probing model vulnerabilities and have been extensively studied \cite{szegedy2014intriguing,zhangsurvey,madry2018towards,carlini2017towards,long2024convergence,zhang2024effective,zhang2025maa}. These attacks typically involve crafting imperceptible perturbations that, when applied to input data, induce incorrect model predictions.

Adversarial attacks are originally developed for unimodal settings, where only a single data modality is considered \cite{szegedy2014intriguing,zhang2021proactive,madry2018towards}. Recent work has explored attacks on specific multimodal tasks, such as image–text retrieval \cite{zhang2023proactive,zhang2021privacy,zhang2021high,zhang2021aggregation} and visual question answering \cite{yin2024vqattack}. These approaches typically rely on task-specific losses or model outputs, limiting their applicability to task-agnostic VLP models. Current VLP attack methods focus on features that are independent of specific tasks and output, i.e., enlarging the feature distance between adversarial data and original data to break cross-modal alignments. However, current methods \cite{zhang2022towards,lu2023set,zhang2025maa} generally customize adversarial perturbations for each sample based on individual characteristics. These sample-wise approaches have to learn adversarial perturbations for new data from scratch, resulting in significant computational overhead. This would significantly deter their applications from large-scale cases. 

Zhou \textit{et al.} \cite{zhou2023advclip,zhang2024universal} make early attempts to learn universal adversarial attacks for the image modality. However, the method in \cite{zhou2023advclip} is restricted to specific models, and neither work exploits text-based attacks, which could further enhance adversarial effectiveness. Fang \textit{et al.} \cite{fang2025one} consider both image and text universal attacks, where the text-side attacks are implemented through word substitutions. These substitutions are obtained by first learning adversarial embeddings and then searching the corpus for words that are semantically aligned with these embeddings. However, this method requires access to a predefined word library, and the mismatch between the learned adversarial embeddings and their final token-level realizations can weaken the attack effectiveness. In addition, current methods are developped in the black-box settings, where the victim model is inaccessible and a surrogate model is leveraged as the target. However, different VLP models are pre-trained with diverse architectures, learning objectives, and datasets, and are further fine-tuned for specific downstream tasks \cite{radford2021learning,li2021align,yang2022vision}. These variations cause different VLP models to rely on distinct visual features or patterns. For instance, image–text retrieval primarily relies on global image–text similarity, while image captioning demands fine-grained visual recognition and semantic reasoning. Existing attacks are prone to overfitting to the surrogate model and failing to transfer to different downstream tasks, models, and data. 

\begin{figure*}
\center
\includegraphics[width=0.9\textwidth]{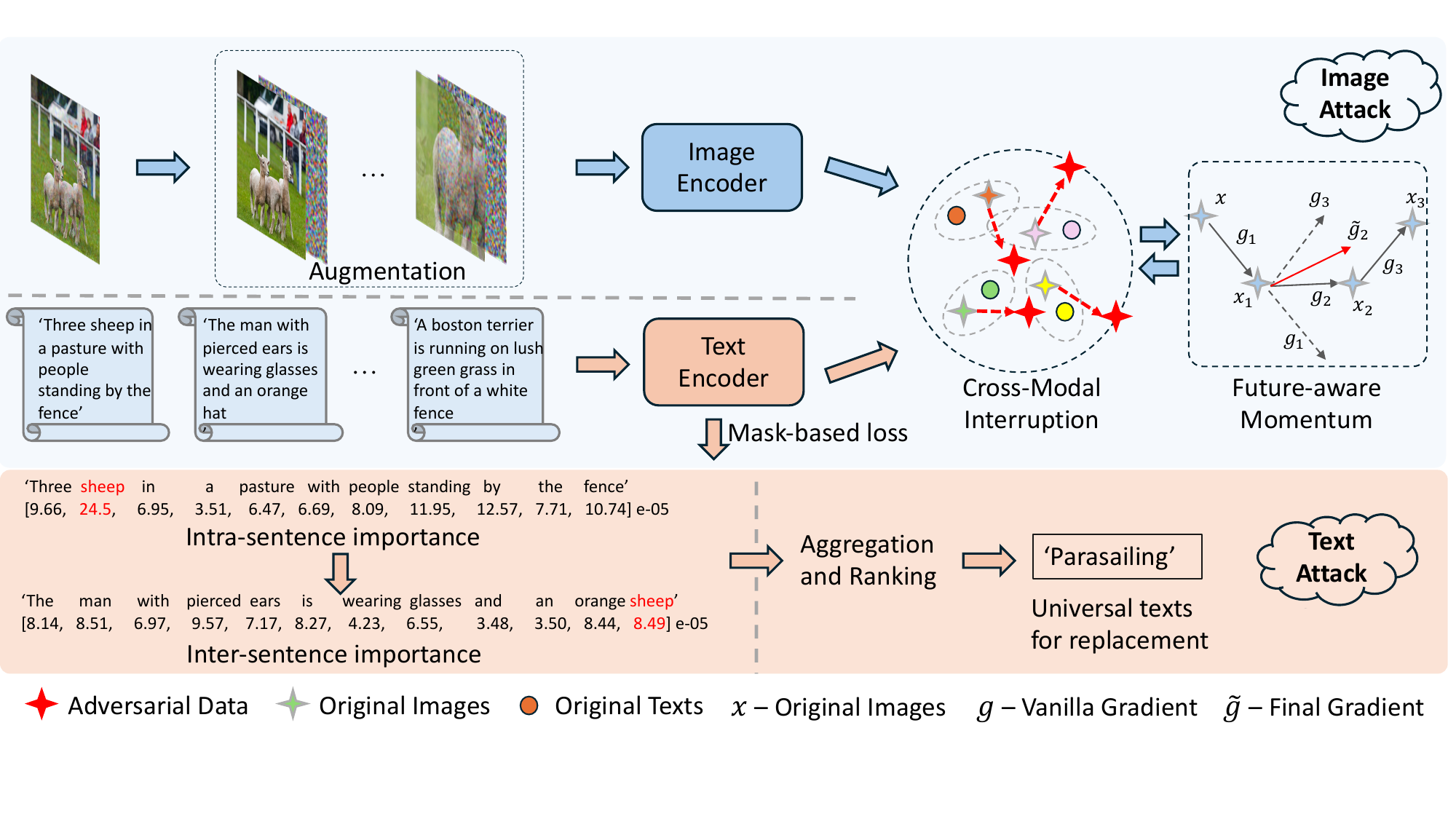}
\caption{An illustration of the proposed HRA framework, which hierarchically refines universal perturbations to enhance transferability. For the image modality, future-aware momentum leverages both historical and predicted future gradients to regularize the current update direction and prevent convergence to local optima. For the text modality, universal perturbations are learned by hierarchically modeling textual importance, including intra- and inter-sentence ranking. Data augmentation is applied to better exploit cross-modal interactions.}
  \label{fig1}
\end{figure*}

Considering this, we propose Hierarchical Refinement Attack (HRA), which enhances the universal adversarial transferability to disrupt cross-modal alignments. Due to the fundamentally different natures of visual and textual data, HRA adopts modality-specific refinement strategies. Specifically, for images, which are continuous, we investigate the optimization trajectory, which is closely related to overfitting. Overfitting often occurs when the optimization process of gradient-based methods converges to local optima. To mitigate this problem, we propose future-aware momentum, which incorporates historical and predicted future gradients to regularize the current update and prevent the optimization from local optima. For text, which is discrete, we implement a simple yet effective adversarial text attack that leverages intra-sentence and inter-sentence importance measures to extract influential words as uniform adversarial perturbations. Extensive experiments demonstrate the effectiveness of the proposed method.

The main contributions of this work are:
\begin{itemize}

\item A novel multi-modal attack method - HRA is proposed, which learns UAPs for both image and text modalities. The generated UAPs can be used for new data, tasks and models without retraining.

\item HRA hierarchically refines UAPs by leveraging past and future gradients to regularize the optimization trajectory, together with intra- and inter-sentence importance measures to enhance adversarial transferability.

\item Thorough experiments are conducted on a wide range of target VLP models in multiple downstream tasks against various datasets. The results demonstrate that the proposed method effectively improves the transferability of universal multimodal attacks.

\end{itemize}

The remainder of the paper is organized as follows: Section \ref{Rel} provides a brief review of related work. Section \ref{METHOD} presents a detailed introduction to our methodology. In Section \ref{Exp}, we conduct extensive experiments, including attacks on various VLP models, datasets, and downstream tasks, as well as ablation studies and visualizations, to demonstrate the effectiveness of the proposed method. Finally, Section \ref{CONCLUSION_future} concludes the paper and discusses current limitations and potential directions for future research.

\section{Related work}\label{Rel}

\subsection{Vision-Language Pre-training}

Digital and realistic worlds encompass diverse data types, e.g., image, text, audio, video, and so on. To better understand the world, vision-language pre-training (VLP) has been developed to understand visual and language content, capture their relations and generate multimodal content. It harnesses vast amounts of unlabeled multi-modal data,  using self-supervised learning techniques to gain representative feature learning ability \cite{li2022blip,radford2021learning,li2021align,yang2022vision,zhang2025step}. VLP models exhibit outstanding generalizability and transferability, allowing adaptation to diverse datasets and downstream tasks through fine-tuning. Representative VLP methods include CLIP \cite{radford2021learning}, BLIP \cite{li2022blip}, ALBEF \cite{li2021align}, and TCL \cite{yang2022vision}. CLIP aligns visual-linguistic modalities by projecting images and texts into a unified feature space, using contrastive learning to maximize similarity for matched pairs and increase the distance for unmatched pairs. BLIP \cite{li2022blip} enhances dataset quality by synthesizing captions for web images, employing three contrastive learning objectives to jointly pre-train the model, i.e., image-text contrastive learning, image-text matching, and image-conditioned language modelling. ALBEF \cite{li2021align} first aligns unimodal representations of pair images and texts, and then fuses them with cross-modal attention to obtain joint representations. TCL \cite{yang2022vision} utilizes contrastive learning to perform both inter- and intra-modal alignment while preserving mutual information between global and local representations in each modality. 

\subsection{Adversarial Attack}
Real-world applications often involve uncertainty, randomness, and potential threats, making it crucial to ensure that developed models are both robust and secure. Adversarial attacks are one of the most effective methods for evaluating model robustness, which aims to induce incorrect predictions of target models by injecting imperceptible perturbations (i.e., adversarial perturbations) into input data \cite{szegedy2014intriguing,zhang2023proactive,zhang2021privacy,zhangsurvey}. According to the scope, existing adversarial perturbations can be roughly categorized into sample-wise \cite{madry2018towards,carlini2017towards}, class-wise \cite{benz2021universal} and universal adversarial perturbations (UAPs) \cite{moosavi2017universal,li2020regional,liu2023enhancing}, which are designed for individual samples, specific classes, and all data, respectively. Generally, sample-wise and class-wise perturbations are more effective than universal perturbations as they can exploit vulnerabilities specific to each sample or class. However, they have to be generated from scratch for new samples or classes, limiting their scalability in large-scale applications. In contrast, universal perturbations are more efficient, as they can be reused across different data once learned. Nevertheless, they fail to leverage data-specific characteristics and are more prone to overfit to target models during training. 

\subsubsection{white-box and black-box attacks} Early adversarial attacks are developed in \textit{white-box} settings, assuming full access to target models, tasks, and data \cite{szegedy2014intriguing,zhang2023proactive,zhang2021privacy}. In practice, such information is often unavailable, and generating perturbations for all possible targets is infeasible. This motivates \textit{black-box} attacks, where target information is inaccessible. Existing approaches mainly fall into \textit{query-based attacks} \cite{papernot2017practical,chen2017zoo,zhu2023efficient} and \textit{transfer-based attacks} \cite{liu2016delving,dong2019evading,wang2021boosting}. 

Query-based attacks infer adversarial perturbations by analyzing model feedback on perturbed inputs and typically estimate gradients via techniques like finite-difference approximations \cite{chen2017zoo}. However, they still require access to the output of target models, limiting practicality. Transfer-based attacks alleviate this constraint by leveraging substitute models, but their effectiveness and transferability is often compromised due to overfitting to substitute models. To handle this, several strategies have been proposed. Data augmentation is employed to increase input diversity \cite{lin2019nesterov,wang2021admix}, while ensemble-based attacks optimize perturbations against multiple surrogate models to improve generalization \cite{xiong2022stochastic,chen2023adaptive}. However, ensemble methods require access to multiple models and incur additional computational overhead. Momentum-based attacks stabilize update directions by accumulating historical gradients, thereby reducing convergence to poor local optima and improving transferability \cite{long2024convergence,inkawhich2019transferable,dong2018boosting}. 

\subsubsection{Adversarial attack against VLP models}
Adversarial attacks against VLP models are still in their early stages, with only a few methods proposed. Compared to conventional multimodal attacks, attacking VLP models is more challenging due to the lack of access to downstream tasks, models and fine-tuning details. Existing methods primarily concentrate on the feature space and learn sample-wise perturbations by enlarging the feature distance between adversarial data and original data \cite{zhang2022towards,lu2023set,zhang2025maa}. To further improve transferability, Lu \textit{et al.} \cite{lu2023set,zhang2025maa} propose data augmentation to create more diverse image-text pairs, aiming to further exploit cross-modal interactions and mitigate overfitting. Despite the progress made, current approaches usually learn sample-wise perturbations, which are tailored to the specific characteristics and vulnerabilities of individual data samples. Consequently, attackers have to learn tailored perturbations for each new data from scratch. This would incur huge computational costs and lead to less scalability for large-scale scenarios. To solve this, Zhou \textit{et al.} \cite{zhou2023advclip} explore the use of universal adversarial perturbations, though their approach is restricted to CLIP. Zhang \textit{et al.} \cite{zhang2024universal} proposes to learn universal adversarial perturbations by enhancing their local utility and leveraging a ScMix data augmentation to promote data diversity to prevent overfitting. However, these methods focus solely on image-modality attacks and neglect the text modality, thereby limiting their effectiveness. Fang \textit{et al.} \cite{fang2025one} present an early attempt to generate both image and text attacks. Their method trains generators to produce adversarial examples by modeling data distributions, which risks overfitting to the source distribution. In addition, the method generates adversarial text embeddings and then searches for semantically similar tokens in the corpus, creating a mismatch between embedding-level optimization and the final token-level perturbations.

\section{PROPOSED METHOD}\label{METHOD}
\subsection{Preliminaries}
The proposed HRA learns universal adversarial perturbations (UAPs) for both image and text modalities in a black-box setting, where target VLP models, datasets, and downstream tasks are inaccessible during training. To this end, we adopt a transfer-based attack framework that leverages a surrogate model and dataset to generate UAPs transferable across diverse models, datasets, and tasks.

Our approach is generic and imposes no assumptions on the surrogate model. For example, some VLP models employ a multimodal encoder \cite{li2021align,yang2022vision}, while others do not \cite{radford2021learning}. To ensure broad generality, we uniformly utilize the unimodal encoders available in each VLP model, i.e., the image encoder and the text encoder, denoted as $f_I$ and $f_T$, respectively. Let $\mathcal{D}_{src} = \{({x}_i, {y}_i)\}_{i=1}^n$ denote a multimodal training dataset, where $({x}_i, {y}_i)$ is an image-text pair and $n$ is the number of pairs. The objective is to learn UAPs that can mislead other VLP models to incorrectly associate images and texts in a target dataset $\mathcal{D}_{tar}$ at the reference time. For the image modality, we learn UAPs $\delta_I$ constrained by an $l_{\infty}$ norm to ensure imperceptibility, i.e., $\|\delta_I\|_{\infty} \leq \epsilon_I$. Here, $\epsilon_I$ represents the perturbation magnitude. For the text modality, due to token discreteness, we implement a universal attack by consistently substituting a word in the text with a replacement word, with imperceptibility controlled by a substitution budget $\epsilon_T$. 

\subsection{Overview}

To enhance adversarial transferability, we propose Hierarchical Refinement Attack (HRA), a universal multimodal adversarial framework. For the image modality, HRA directly regularizes the optimization trajectory, as overfitting arises when updates converge to local optima. By incorporating both past and predicted future gradients, it stabilizes update directions and prevents premature convergence. For the text modality, HRA hierarchically ranks word importance by masking individual tokens and measuring the divergence between masked and original representations, accounting for both intra- and inter-sentence contexts. An illustration of HRA is shown in Fig. \ref{fig1}.

\subsection{Future-aware Momentum} 
VLP models commonly bridge the gap between heterogeneous data, i.e., images and texts, by creating a shared representation space in which semantically related visual and linguistic inputs are closely aligned. This makes it natural for an attacker to disturb the alignment between paired images and texts. Specifically, given a surrogate model, A typical method for learning image UAPs is to enlarge the feature distance between matched images and texts: 
\begin{equation}\label{eq1-1}
\begin{split}
& \underset{\delta_I}{\arg\max } \mathcal{L} \left({x}, y, \delta_I \right) = \sum _{i=1}^{n} \left(\ell \left(f_I\left({x}_i + \delta_I
\right),f_I\left({x}_i\right) \right) \right.\\ 
& \left. \ \ \ \ \ \ \ \ \ \ \ \ \ \ \ \ \ \ \ \ \ \ \ \ \ \ \ \ \ + \ell \left(f_I\left({x}_i + \delta_I\right), f_T\left({y_i}\right) \right)  \right),\\
\end{split}
\end{equation}
where $\ell$ is a loss function to quantify the difference between representations of two samples, e.g., the KL-divergence loss.

Iterative gradient-based optimization methods such as PGD \cite{kurakin2016adversarial,madry2018towards} are commonly used to solve the above objective. However, under multi-round optimization, the update process tends to converge to suboptimal local minima, leading to overfitting \cite{dong2018boosting,wang2021boosting,wang2021enhancing,lin2019nesterov}. Momentum-based methods have been shown to stabilize update directions by accumulating historical gradients \cite{long2024convergence,inkawhich2019transferable}. Classical momentum regularizes the optimization trajectory solely based on past gradients. In the context of universal adversarial perturbation learning, where gradients are aggregated across the entire dataset, historical gradients may become misaligned with the current optimization direction, limiting attack effectiveness. Moreover, conventional momentum operates within a restricted temporal scope, resulting in a narrow exploration space. To address these limitations, we propose hierarchical future-aware momentum, which additionally incorporates estimated future gradients to regularize the optimization trajectory. This expanded temporal hierarchy stabilizes updates, broadens the search space, and effectively mitigates overfitting.

Specifically, denote the current mini-batch as $(x^{m_I}, y^{m_I})$, where ${m_I}$ is the training step. The gradient of the current optimization step is:
\begin{equation}\label{eq2-1}
\begin{split}
&  {g}^{m_I} \leftarrow \frac{1}{\left|x^{m_I}\right|} \nabla_{\delta_I^{m_I}} \mathcal{L} \left(x^{m_I}, y^{m_I}, \delta_I^{m_I} \right),\\
\end{split}
\end{equation}

We also record the gradient of the last step:
\begin{equation}\label{eq2-2}
\begin{split}
&  {g}_p^{m_I-1} \leftarrow \frac{1}{\left|x^{m_I-1}\right|} \nabla_{\delta_I^{m_I-1}} \mathcal{L} \left(x^{m_I-1}, y^{m_I-1}, \delta_I^{m_I-1} \right),\\
\end{split}
\end{equation}

The future gradient is calculated by:
\begin{equation}\label{eq2-3}
\begin{split}
& {g}_f^{m_I} = \frac{1}{d} \sum _{i=1}^{d} {g}^d_f,\\
& s.t. \ {g}^{m_I,d}_f \leftarrow \frac{1}{\left|x^{m_I}\right|} \nabla_{\delta_I} \mathcal{L}_1 \left(x^{m_I}, y^{m_I}, \delta_I^{m_I+d} \right),\\
\end{split}
\end{equation}
Where ${g}_f$ denotes the mean gradient over the next $d$-step. 

We regularize the current gradient using both the previous and future gradients as follows:
\begin{equation}\label{eq2-4}
\begin{split}
&  \tilde{g}^{m_I} \leftarrow {g}^{m_I} + \gamma_1 \cdot {g}^{m_I}_p + \gamma_2 \cdot {g}^{m_I,d}_f,\\
\end{split}
\end{equation}
where $\gamma_1, \gamma_2 \in [0,1)$ are balancing parameters.

To further exploit cross-modal interactions, we augment the dataset following existing practices \cite{lu2023set,zhang2024universal}.

\subsection{Text Attack} 
Different from images, text is discrete in nature and highly sensitive to token replacements, which makes it impossible to directly learn universal text perturbations in the same way as image perturbations. To address this challenge, we propose a simple yet effective text attack based on word substitution using universal trigger words. Our universal text attack does not rely on any external resources. Instead, it identifies optimal substitution tokens directly from the training corpus.

To identify influential words for text attacks, the method first measures the importance of each token in every training sample (intra-sentence importance measures) and aggregates these scores over the entire dataset (inter-sentence importance measures). The importance of a word is computed by masking it in the input text and evaluating the semantic discrepancy between the masked representation and that of the original image-text pair:
\begin{equation}\label{eq3-1}
\begin{split}
& S(w) = \ell (f_I(\hat{y}), f_T({y})) + \ell (f_I(\hat{y}), f_I({x})),\\
\end{split}
\end{equation}
where $w$ denotes a word. ${y}$ and $\hat{y}$ are the original text and its masked counterpart, respectively.

Specifically, the importance of each word in the original sample is first measured and the top-$k$ most salient words in each sample are selected as candidates. Their global influence is then further assessed by randomly substituting one of these candidate words into other sentences and measuring the resulting semantic shift. Universal trigger words are finally obtained by ranking all candidate tokens according to their aggregated influence scores and choosing the highest-ranked ones. To preserve imperceptibility, we restrict the text perturbation budget to $\epsilon_T =1$, allowing only one token substitution. At test time, we consider two attack variants: HRA$_{ran}$, which randomly replaces a word, and HRA$_{imp}$, which replaces the most important word as in \cite{fang2025one}.

The detailed algorithm is summarized in \textbf{Algorithm \ref{alg1}}. 

\begin{algorithm}[t!]
       \caption{Hierarchical Refinement Attack}
       \label{alg1} 
       \textbf{Require}: Training data $\mathcal{D}$, mini-batch size $l$, iteration times $M_I, M_T$,  parameters $\epsilon_I, \epsilon_T, \alpha, \beta_1, \beta_2, \gamma_1, \gamma_2, d$; \\
        \textbf{Require}: Randomly initialize $\delta_I$, initialize ${g}_p^{0} = 0$;\\
        // Image attack;\\
        // Exploit diverse matching captions for each image to augment the dataset;\\
        \For{$m_I = 1 \rightarrow M_I$}{
        \For{$(x^{m_I}, y^{m_I}) \sim \mathcal{D}$}{
            $(\tilde{x}^{m_I}, y^{m_I}) \leftarrow (x^{m_I}, y^{m_I})$ //  Augment each image-text pair;\\
            ${g}^{m_I} \leftarrow \frac{1}{\left|x^{m_I}\right|} \nabla_{\delta^{m_I}_I} \mathcal{L} \left(x^{m_I}, y^{m_I}, \delta_I^{m_I} \right)$ // Calculate the current gradient according to Eq.(\ref{eq2-1});\\
            ${g}^{m_I}_f \leftarrow \frac{1}{\left|x^{m_I}\right|} \nabla_{\delta^{m_I}_I} \mathcal{L}_1 \left(x^{m_I}, y^{m_I}, \delta_I^{m_I,d} \right)$ // Calculate future gradients according to Eq.(\ref{eq2-3});\\
            $\tilde{g}^{m_I} \leftarrow {g}^{m_I} + \gamma_1 \cdot {g}_p^{m_I-1} - \gamma_2 \cdot {g}_f^{m_I}$  // Update the gradient according to Eq.(\ref{eq2-4});\\
            $ \delta_I^{m_I} \leftarrow \operatorname{Clip}_{\epsilon_I}(\delta_I^{m_I} +\alpha \cdot \operatorname{sign}(\tilde{g}^{m_I}))$ //Update image UAPs;\\
        }
        ${g}_p^{m_I} = \tilde{g}^{m_I}$ // Update the previous gradient;\\
        }
        // Text attack;\\
        \For{$m_T = 1 \rightarrow M_T$}{
        \For{$(x^{m_T}, y^{m_T}) \sim \mathcal{D}$}{
        $S(w) = \ell (f_I(\hat{y^{m_T}}), f_T({y^{m_T}}) )$ // Measure and record the importance of each word in a text sample according to Eq.(\ref{eq3-1});\\
        }}
        $\delta_T \leftarrow Rank(S)$ // Rank the importance and select the important influential word as the perturbation;\\
        return $\delta_I, \delta_T$\;
\end{algorithm}

\section{Experiments}\label{Exp}

\begin{table*}[!t]
\center
\caption{The attack success rate $(\%)$ of R@1 in image-text retrieval. Grey cells denote white-box attack results. \textbf{Bold} indicates the best performance, and \underline{*} marks the second best.}
\label{tab_cross}
\centering
\scalebox{0.95}{
\begin{tabular}{c|c|cc|cc|cc|cc|cc|cc|cc}
\toprule
\multicolumn{2}{c}{\multirow{1}{*}{\textbf{Test Dataset}}}  & \multicolumn{12}{c}{\textbf{Flickr30K}} \\ \hline 

\multicolumn{2}{c}{\multirow{3}{*}{\textbf{Target Model}}} & \multicolumn{10}{c}{\textbf{CLIP}} & \multicolumn{2}{c}{\multirow{3}{*}{\textbf{ALBEF}}} & \multicolumn{2}{c}{\multirow{3}{*}{\textbf{TCL}}}\\
\cmidrule(lr){3-12} 
\multicolumn{2}{c}{}& \multicolumn{2}{c}{\text{ResNet50}}  &\multicolumn{2}{c}{\text{ResNet101}}  & \multicolumn{2}{c}{$\text{ViT-B/16}$}   & \multicolumn{2}{c}{$\text{ViT-B/32}$}   & \multicolumn{2}{c}{$\text{ViT-L/14}$}  &\multicolumn{2}{c}{} & \multicolumn{2}{c}{} \\ \hline

\multicolumn{1}{c|}{Source Model} & {Method} & I2T & T2I & I2T & T2I  & I2T & T2I & I2T & T2I & I2T & T2I & I2T & T2I & I2T & T2I \\ \hline

\multirow{7}{*}{CLIP$_\text{ViT-B/16}$}  
&{AdvCLIP}      &{19.12}  & {{27.58}} &{15.33}    &17.29  &\cellcolor[HTML]{EFEFEF}{76.07} & \cellcolor[HTML]{EFEFEF}{76.58 }   &{15.83}    & 24.29  & {11.58}    & {17.85}   &{3.96}& {7.83} & {18.02}& {20.67}\\
{}  &{SGA}   &{38.44}  & {{48.23}} &{33.84}    &38.35  &\cellcolor[HTML]{EFEFEF}{\textbf{92.39}} & \cellcolor[HTML]{EFEFEF}{93.78}   &{20.49}    & 32.22  & {17.79}    & {24.03}   &{8.45}& {12.51} & {10.01}& {15.41}\\
{} & {ETU}   &{56.83}  & {{61.27}} &{52.49}    &54.27  &\cellcolor[HTML]{EFEFEF}{88.47} & \cellcolor[HTML]{EFEFEF}{92.69}   &{22.58}    & 33.70  & {20.86}    & {28.54}   &{13.14}& {17.28} & {18.55}& {21.57}\\
{}  & {FD-UAP}   &{38.06}  & {{46.83}} &{30.27}    &36.33  &\cellcolor[HTML]{EFEFEF}{85.89} & \cellcolor[HTML]{EFEFEF}{83.31}   &{20.61}    & 31.83  & {16.07}    & {24.39}   &{8.55}& {13.89} & {10.96}& {15.98}\\
{} & C-PGC    &{67.43}  & {{74.58}} &{59.13}    &70.12  &\cellcolor[HTML]{EFEFEF}{87.98} & \cellcolor[HTML]{EFEFEF}{89.69}   &\underline{48.59}    & 62.31  & {37.30}    & {57.60}   &{25.23}& {35.90} & \underline{43.84}& {59.33}\\

{} & {HRA$_{ran}$}  &\underline{70.50}  & \underline{84.46} &\underline{71.52}    &\underline{85.08}  &\cellcolor[HTML]{EFEFEF}{86.75} & \cellcolor[HTML]{EFEFEF}{\underline{94.14}}   &{45.15}    & \underline{68.91}  & \underline{54.11}    & \underline{79.77}   &\underline{32.22}& \underline{57.65} & {43.52}& \underline{62.57}\\

{} & {HRA$_{imp}$}  &{\textbf{76.63}}  & {\textbf{88.54}} &{\textbf{77.39}}    &\textbf{88.99}  &\cellcolor[HTML]{EFEFEF}{\underline{90.55}} & \cellcolor[HTML]{EFEFEF}{\textbf{95.72}}   &{\textbf{54.97}}    & \textbf{79.35}  & {\textbf{61.84}}    & {\textbf{85.73}}   &{\textbf{40.98}}& {\textbf{70.28}} & {\textbf{52.37}}& {\textbf{76.95}}\\

\hline

\multirow{7}{*}{CLIP$_\text{ResNet50}$}  
&{AdvCLIP}     &\cellcolor[HTML]{EFEFEF}{78.77} &\cellcolor[HTML]{EFEFEF}{82.64}  & {{22.99}} &{27.72}    &6.01  &{13.85} & {15.21 }   &{24.36}    & 10.06  & {15.59}    & {3.55}   & {{7.46}} & {17.61}& {20.69}\\
{}  &{SGA}   &\cellcolor[HTML]{EFEFEF}{\textbf{98.16}} &\cellcolor[HTML]{EFEFEF}{\textbf{98.68}}  &27.71 &36.43   &10.92    &{16.33} & {17.55}   &25.45    &10.31  & {19.07}    & 5.74   & 9.92 & {6.85}& {12.67} \\
{} & {ETU}   &\cellcolor[HTML]{EFEFEF}{\underline{95.79}} &\cellcolor[HTML]{EFEFEF}{\underline{98.63}}  & {{37.04}} &{44.94}    &9.33     &{17.49} & {16.58 }   &{27.74}    & 9.82  & {18.65}    & {5.42}   & {{10.59}} & {18.55}& {{21.57}} \\
{}  & {FD-UAP}   &\cellcolor[HTML]{EFEFEF}{78.80} &\cellcolor[HTML]{EFEFEF}{89.81}  &12.90 &19.45    &6.01     &12.27 & 14.85   &23.36    & 7.61  & 15.69    & 4.90  &8.35 &6.11 &11.33 \\
{} & C-PGC   &\cellcolor[HTML]{EFEFEF}{92.34} &\cellcolor[HTML]{EFEFEF}{91.77}  &\underline{68.45} &76.47    &\underline{31.17}  &\underline{48.87} &\underline{40.25}   &\underline{59.18}    &\underline{30.06}  &48.36    &17.52  &34.14 &\underline{22.76} &\underline{42.38} \\

{} & {HRA$_{ran}$}   &\cellcolor[HTML]{EFEFEF}{94.76} &\cellcolor[HTML]{EFEFEF}{96.57}  & {63.60} &\underline{77.29}    &{29.20}  &{48.61} & {32.15}   &{51.80}    & {25.15}  & \underline{51.97}  & \underline{19.19}   & \underline{36.13} & {21.29} &\textbf{36.88}  \\

{} & {HRA$_{imp}$}    &\cellcolor[HTML]{EFEFEF}{95.27} &\cellcolor[HTML]{EFEFEF}{97.26}  & \textbf{79.82} &\textbf{87.24}    &\textbf{37.67}  &\textbf{62.24} & \textbf{42.33}   &\textbf{65.59}    & \textbf{33.25}  & \textbf{67.11}  & \textbf{30.03}   & \textbf{53.95} & \textbf{34.14} &\textbf{54.76}  \\
\hline

\multirow{7}{*}{ALBEF} 
&{AdvCLIP}  & {19.41}  & 27.62  & {13.67}  & 17.22  & 8.96    &15.34 & 17.18  & 25.35     & 9.94   & {17.46}    & \cellcolor[HTML]{EFEFEF} {63.19}    & \cellcolor[HTML]{EFEFEF}{63.38}  & 37.72 & {36.76} \\
{}  &{SGA} & 18.14  & 25.28  & {11.37} & 17.26  & 6.99   &12.79 & 16.69  & 23.71  & 8.10   & {15.59}      & \cellcolor[HTML]{EFEFEF}{74.87}   & \cellcolor[HTML]{EFEFEF}{72.99}   & 18.65& {19.62} \\
{}  &{ETU}  & {26.56}  & {35.03} & {20.31} & {25.08} & {10.06}   &{17.72}  & {17.18}  & {26.22}   & {13.01}    & {21.55}      & \cellcolor[HTML]{EFEFEF}{83.00}   & \cellcolor[HTML]{EFEFEF}{87.07}   & {32.98}& {31.50}\\
{}  & {FD-UAP} & {20.43}  & {30.19} & {12.90} &{19.01} &8.83   &14.66  & 17.55  &25.29   & 9.94    & 17.91     & \cellcolor[HTML]{EFEFEF}{{67.47}}   & \cellcolor[HTML]{EFEFEF}{63.49}   & 27.82 &25.69\\
{} & C-PGC  &44.32  &55.92 &\underline{36.02} &50.94 &29.69   &42.14  &\underline{39.75}  &\underline{53.99}   &30.55   &45.46     & \cellcolor[HTML]{EFEFEF}{84.36}   & \cellcolor[HTML]{EFEFEF}{81.10}   &43.94 &50.57\\

{} & {HRA$_{ran}$}  & \underline{53.26}  & \underline{66.93} & {\underline{36.02}} & \underline{53.00} & \underline{30.92}   &\underline{48.97}  & {33.87}  & {46.07}   & \underline{32.64}    & {\underline{49.71}}      & \cellcolor[HTML]{EFEFEF}{\underline{88.01}}   & \cellcolor[HTML]{EFEFEF}{\underline{88.38}}   & \underline{55.85}& \underline{56.71}\\

{} & {HRA$_{imp}$}  & \textbf{60.28}  & \textbf{75.71} & {\textbf{52.87}} & \textbf{69.98} &\textbf{40.98}  & \textbf{60.02}  & \textbf{46.26}  & \textbf{59.57}   & \textbf{41.10}  &\textbf{60.89}  & \cellcolor[HTML]{EFEFEF}{91.76}   & \cellcolor[HTML]{EFEFEF}{92.42}   & \textbf{65.12}& \textbf{69.24}\\\hline

\multirow{7}{*}{TCL} 
{}  &{AdvCLIP} & 24.27   & 35.68& {17.62} & 22.74  & 8.22   &16.33  & 17.18   & 28.16  & 11.41    & {19.78}     & 23.15   & {24.58}    & \cellcolor[HTML]{EFEFEF}{60.17}& \cellcolor[HTML]{EFEFEF}{{54.12}} \\
{}  &{SGA}  & {20.31}  & 29.16 & {13.28}  &19.38 & 7.36  & {13.66} & 16.97  & 25.19     & 9.45  & {18.65}   & 18.87   & {20.65}   & \cellcolor[HTML]{EFEFEF}{83.63} & \cellcolor[HTML]{EFEFEF}{86.37} \\
{} &{ETU}   & {27.59} & {39.69} & {20.82}  & {26.96}& {9.20}  & {17.94}   & {17.44}    & {28.61}   & {12.64}   & {20.59}   &{23.25}   & {25.44}    &\cellcolor[HTML]{EFEFEF}{\underline{94.10}} & \cellcolor[HTML]{EFEFEF}{87.60} \\
{}  & {FD-UAP}  & 18.14 & 27.92 &10.98  & 17.22& 7.85  & 14.50   & 16.07    & 25.26   & 9.20   &17.69    &13.56   & 18.26    &\cellcolor[HTML]{EFEFEF}{{67.47}} & \cellcolor[HTML]{EFEFEF}{63.49} \\
{} & C-PGC  &48.15  &60.03 &38.19 &53.93 &32.68  &45.65  &\underline{38.90}  &51.90   &\underline{34.48}   &49.55     &37.02    & 42.26   &\cellcolor[HTML]{EFEFEF}{85.88} &\cellcolor[HTML]{EFEFEF}{76.10}\\

{} & {HRA$_{ran}$}  & {\underline{52.36}}  & \underline{72.42} & {\underline{52.23}}  & \underline{70.67}& {\underline{32.76}}  & \underline{65.59}   & {36.20}    & \underline{62.27}    &{32.27}   & \underline{63.79} &\underline{53.28}   & \underline{63.24}  &\cellcolor[HTML]{EFEFEF}{91.15} & \cellcolor[HTML]{EFEFEF}{\underline{90.98}} \\

{} & {HRA$_{imp}$}  & \textbf{63.22}  & \textbf{79.93} & {\textbf{58.24}} & \textbf{81.30} &\textbf{41.84}  & \textbf{76.97}  & \textbf{47.36}  & \textbf{73.90}   & \textbf{42.58}  &\textbf{76.03}  & {64.13}   & {77.03}   & \cellcolor[HTML]{EFEFEF}\textbf{94.20}& \cellcolor[HTML]{EFEFEF}\textbf{94.38} \\ 
\bottomrule 
\end{tabular}}
\end{table*}

\subsection{Settings}
\noindent\textbf{Downstream tasks and datasets.} 
To comprehensively evaluate the performance of the proposed method, we conduct experiments on four vision-language tasks with three datasets. 

Tasks include the image-text retrieval, the image captioning, and the visual grounding. The image-text retrieval aims to find relevant data from a modality based on a query from another modality. This task consider the ranking between different modalities. Image captioning endeavours to generate a descriptive textual caption for a given image, which requires a model to capture details of visual data. Visual grounding is to associate textual descriptions with specific regions or objects in an image, which requires VLP models to understand and localize objects based on textual input. 

Three widely used datasets, i.e., Flickr30K \cite{plummer2015flickr30k}, MSCOCO \cite{lin2014microsoft} and RefCOCO+ \cite{yu2016modeling}, are selected. Flickr30K consists of 31,783 images, each accompanied by five descriptive captions. The MSCOCO 2014 dataset contains 164K images, each annotated with approximately five captions. RefCOCO+ includes 141,564 referring expressions corresponding to 49,856 objects across 19,992 images.

\noindent\textbf{Models.} Four widely-used VLP models are utilized to test the proposed method, i.e.,  CLIP \cite{radford2021learning}, ALBEF \cite{li2021align}, TCL \cite{yang2022vision} and BLIP \cite{li2022blip}.  For CLIP, different image encoders are utilized, including vision transformers (i.e., ViT-B/16,  ViT-B/32, and  ViT-L/14\cite{dosovitskiy2020image}) and CNNs (i.e., ResNet50, and ResNet101 \cite{he2016deep}). The text encoder is a 6-layer transformer. For BLIP, we choose the one that consists of a ViT-B/16 and a 6-layer transformer as the image and text encoder to attack. In addition, BLIP is mainly used for experiments on the image captioning task. ALBEF and TCL take ViT-B/16 as the image encoder and adopt a 6-layer transformer for both the text encoder and multimodal encoder. 

\noindent\textbf{Evaluation metric.} For image-text retrieval, we utilize the Attack Success Rate (ASR) as a metric to quantify the effectiveness of the proposed attack and all compared baselines. ASR is calculated as the percentage of adversarial examples that successfully deceive the model, providing a reliable measure of the attackers' effectiveness. For other tasks, we compare the performance of VLP models on adversarial examples with their performance on the original examples.

\noindent\textbf{Implementation details.} For the fundamental experiments, we set the perturbation magnitudes to $\epsilon_I =12/255$ for images and $\epsilon_T=1$ for texts. We further evaluate the proposed method under varying perturbation budgets. PGD is utilized to solve the optimization problem for the image attack, with the number of iterations $M_I =100$ and the step size as $\alpha = \epsilon_I/M_I*1.25$. The batch size is set as 16. $\beta_1=0.8, \beta_2=0.2, \gamma_1 =0.9, \gamma_2=0.1, M_T =15$. For $\ell$, we employ the KL-divergence to quantify the discrepancy between two samples. We follow the data augmentation strategy introduced in \cite{zhang2024universal}.

\begin{table}[!t]
\caption{Attack performance on visual grounding. The training and test dataset are Flickr30K and RefCOCO+, respectively. ALBEF built for visual grounding is used as the target model. The ``Baseline'' denotes the performance of the target model on the original data. Lower values represent better adversarial transferability. \textbf{Bold} indicates the best performance, and \underline{*} marks the second best.}
\label{tab_trans_3}
\centering
\scalebox{1}{
\begin{tabular}{@{}c|ccc|cccc@{}}
\toprule
\multicolumn{1}{c}{{Source Model}}  & \multicolumn{3}{c}{{CLIP$_\text{ViT-B/16}$}}   & \multicolumn{3}{c}{ALBEF}  \\ \midrule
{} & {Val} & {TestA}   & {TestB}   & {Val} & {TestA}   & {TestB} \\ \hline
{Baseline}   & {51.24}  &{56.71} &{44.79}     & {51.24}  &{56.71} &{44.79}   \\
{AdvCLIP}  &{50.45}  &{54.49} & {43.32}  &{48.21}  &{52.48}  &{40.99}  \\
{SGA}   &{49.43}  &{53.95}  & {42.46}   &{47.64}  &{52.79}  &{38.58} \\
{ETU}  &{48.50}  &{51.54}  & {41.66}  &{45.49}  &{49.70}  &{38.00} \\
{FD-UAP}   &{49.09}  &{52.74}  & {42.28}  &{46.08}  &{51.36}  &{36.53} \\
C-PGC   &41.59  &44.22  &36.74  &39.22  &42.82  &33.83 \\
{HRA$_{ran}$}  &{\underline{34.21}}  &{\underline{34.93}}  &{\underline{31.85}} & \underline{32.60} &\underline{32.80}  &\underline{28.00}  \\
{HRA$_{imp}$}  &{\textbf{32.85}}  &\textbf{33.86}  &\textbf{31.74}  &\textbf{31.58}  &\textbf{32.01}  & \textbf{28.37}\\
\bottomrule
\end{tabular}}
\end{table}

\begin{table*}[!t]
\caption{Attack performance on image captioning. The training dataset and test dataset are Flickr30K and MSCOCO, respectively. BLIP is taken as the target model. ``Baseline'' denotes the performance of the target model on original data. Lower values represent better adversarial transferability. \textbf{Bold} indicates the best performance, and \underline{*} marks the second best.}
\label{tab_trans_4}
\centering
\scalebox{0.95}{
\begin{tabular}{@{}c|ccccc|ccccccc@{}}
\toprule  
\multicolumn{1}{c}{{Source Model}}  & \multicolumn{5}{c}{{CLIP$_\text{ViT-B/16}$}}  & \multicolumn{5}{c}{{ALBEF}}  \\\midrule
{} & B@4  & {METEOR}   & {ROUGE$\_\text{L}$}  & {CIDEr} & {SPICE}  & B@4  & {METEOR}   & {ROUGE$\_\text{L}$}  & {CIDEr} & {SPICE} \\ \hline
{Baseline}   & 39.31  &30.69 &59.62   &131.38 &23.52 & 39.31  &30.69 &59.62   &131.38 &23.52  \\
{AdvCLIP} & 38.20  &30.03 &58.70   &127.14 &22.88  & 36.59  &29.15 &57.35   &121.79 &21.95  \\
{SGA} & 36.10  &28.89 &57.25   &120.47 &21.85  & 33.92  &27.67 &55.41   &112.28 &20.62   \\
{ETU}  & \underline{34.83} &\underline{28.07} &\underline{56.29}   &\underline{114.81} &\underline{21.02}  & \underline{33.42} &\underline{27.28} &\underline{54.96}   &\underline{109.34} &\textbf{20.11} \\
{FD-UAP}  & {35.91} &{28.78} &{57.15}   &{119.63} &{21.65}  & {34.37} &{28.09} &{56.02}   &{113.26} &{20.86} \\
C-PGC &35.63 &{28.44} &{56.75}   &{117.33} &{21.38}  &33.97 &27.73 &55.50   &112.69 &20.50 \\
{HRA}  & \textbf{32.41} &\textbf{26.80} &\textbf{54.31}   &\textbf{106.60} &\textbf{19.61}  & \textbf{32.67} &\textbf{27.21} &\textbf{54.62}   &\textbf{108.19} &\underline{20.15} \\
\bottomrule 
\end{tabular}}
\end{table*}

\noindent\textbf{Baselines.} We compare our approach with several state-of-the-art methods, including AdvCLIP \cite{zhou2023advclip}, SGA \cite{liu2023enhancing}, ETU \cite{zhang2024universal}, FD-UAP \cite{wang2024improving}, C-PGC \cite{fang2025one}. AdvCLIP, SGA, ETU, and FD-UAP focus solely on image-modality attacks, whereas C-PGC support attacks on both the image and text modalities. 

\noindent\textbf{Evaluation Preparation.} We assess the transferability of the proposed method by launching attacks against a range of VLP models, i.e., CLIP models with different backbones, ALBEF and TCL, in black-box settings. Given that different VLP models may accept inputs of varying sizes, the learned UAPs are resized accordingly before initiating attacks. For instance, UAPs learned on CLIP are resized from $224 \times 224$ to $384 \times 384$ for attacks on ALBEF/TCL, whereas UAPs from ALBEF/TCL to CLIP are resized to $224 \times 224$.

\subsection{Cross-model Transferability}

\begin{table*}[!t]
\center
\caption{The attack success rate $(\%)$ of R@1 under transfer attack from visual grounding on Refcoco+ to image-text retrieval. \textbf{Bold} indicates the best performance, and \underline{*} marks the second best.}
\label{tab_trans_5}
\centering
\scalebox{0.95}{
\begin{tabular}{c|c|cc|cc|cc|cc|cc|cc|cc}
\toprule

\multicolumn{2}{c}{\multirow{1}{*}{\textbf{Test Dataset}}}  & \multicolumn{12}{c}{\textbf{Flickr30K}} \\ \hline 
\multicolumn{2}{c}{\multirow{2}{*}{\textbf{Target Model}}} & \multicolumn{10}{c}{\textbf{CLIP}} & \multicolumn{2}{c}{\multirow{2}{*}{\textbf{ALBEF}}} & \multicolumn{2}{c}{\multirow{2}{*}{\textbf{TCL}}}\\
\cmidrule(lr){3-12} 
\multicolumn{2}{c}{}& \multicolumn{2}{c}{\text{ResNet50}}  &\multicolumn{2}{c}{\text{ResNet101}}  & \multicolumn{2}{c}{$\text{ViT-B/16}$}   & \multicolumn{2}{c}{$\text{ViT-B/32}$}   & \multicolumn{2}{c}{$\text{ViT-L/14}$}  &\multicolumn{2}{c} {} & \multicolumn{2}{c}{} \\

 \hline

\multicolumn{1}{c|}{Source Model} & { Method} & I2T & T2I & I2T & T2I  & I2T & T2I & I2T & T2I & I2T & T2I & I2T & T2I & I2T & T2I \\ \hline
{}  &{AdvCLIP}  &11.88  & 19.52  & 0.26 &0.38  &0.12    &0.48 & 13.13  &20.36   & 8.10    &15.01     & 2.29  &4.58   & 4.03 &7.31 \\
{}  &{SGA} &16.86  &26.48  & 13.03 &17.56  &6.26   &11.98 & 30.83  & 41.35    &7.98   & 16.27   & 8.09  &11.74     & 3.27 & 6.71 \\
{}   &{ETU}   & 20.31 & 28.68 & 14.56 & 19.55  &10.80 &16.49   &19.02  &28.12   & 11.41  &18.91   & 15.95   & 28.74  & 11.49 & 15.00  \\ 
 {}  & {FD-UAP}   & 18.26 &28.27 & 11.75  & 17.29  & 6.13 & 12.31   &16.93   &23.97    &8.34   &15.82   &30.39   & 49.60  & 10.12 & 14.21 \\ 
 {}   & C-PGC  &{35.55} &\underline{51.89} &{31.14}  &\underline{47.74} &{21.31}   &{31.00}   &{22.07}    &{49.44}   &{18.42} &{34.10}   &28.47   &53.11  &27.61 &40.83 \\ 
 {}  & {HRA$_{ran}$}  & \underline{36.40} & {48.44} & \underline{31.16}  &{44.94}  &\underline{22.33} & \underline{33.15}   & \underline{26.01}     & \underline{41.66}    & \underline{20.12}   & \underline{36.28}   &\underline{37.64}   & \underline{55.52}  &\underline{31.09} &\underline{42.71} \\
\multirow{-7}{*}{\tabincell{c}{ALBEF}}   & {HRA$_{imp}$}  &\textbf{45.98} & \textbf{62.92} &\textbf{40.49}  & \textbf{59.83}  & {\textbf{28.83}} & \textbf{46.81}   & \textbf{35.83}     & \textbf{55.12}    & \textbf{30.67}   & \textbf{50.55}   & \textbf{46.51}  & \textbf{66.46}  &\textbf{41.31} &\textbf{57.71} \\

\bottomrule 
\end{tabular}}
\end{table*}
Table \ref{tab_cross} summarizes the attack performance on both image-to-text and text-to-image retrieval tasks, using various source models and the Flickr30K dataset for training. From these results, we draw the following observations.


\subsubsection{Comparison with baselines} All methods exhibit substantial performance degradation when transferred to unseen models and datasets, confirming that different models rely on distinct feature representations and thus incur a significant model gap. From the image-modality perspective, AdvCLIP generates UAPs by destroying neighbourhood relationships in the source models, but fails to account for this in the target model. FD-UAP focuses on low-level features, assuming them to be more transferable. However, this would also lead to overfitting to low-level features specific to the source model. C-PGC produces both image and textual UAPs conditioned on the training distribution. This makes it rely on model-dependent features, limiting its ability to transfer to unseen models. SGA attempts to alleviate overfitting by aggregating gradients from other mini-batches. However, aggregating gradients from many mini-batches may dilute the relevance of the update direction, which explains its limited improvements, particularly on CLIP$_\text{ViT-L/14}$ and ALBEF in Table \ref{tab_cross}. In contrast, our method utilizes both historical and estimated future gradients to regularize the current gradient, providing more informative optimization guidance and reducing the risk of convergence to local optima. Consequently, it achieves superior transferability across models.

\subsubsection{Effects of multimodal attacks} Incorporating text attacks further enhances overall performance, as reflected by the gains of C-PGC and the proposed HRA over other methods. Replacing the most salient words provides additional improvement, evident from the superior results of {HRA${imp}$} compared with {HRA${ran}$}. Interestingly, although C-PGC also replaces important words, {HRA${ran}$} still outperforms it in many cases, highlighting the effectiveness of directly discovering substitution words rather than learning universal embeddings and mapping them back to tokens. Moreover, our method does not rely on a predefined word library, demonstrating its practicality. While identifying the most important words incurs additional computational overhead ({HRA$_{imp}$}), {HRA$_{ran}$} avoids this cost.

\subsubsection{Effects of learning objectives}
Learning objectives also affect transferability. Models trained with similar objectives generally exhibit stronger adversarial transfer, as observed across different CLIP variants and other architectures. For example, UAPs generated on CLIP$_\text{ViT-B/16}$ transfer more effectively to CLIP$_\text{ResNet50}$ than to ALBEF or TCL.

\subsubsection{Asymmetry of attacks} Interestingly, adversarial transferability is asymmetric. strong transfer in one direction does not necessarily imply strong transfer in the opposite direction, e.g., from CLIP$_\text{ViT-B/16}$ to CLIP$_\text{ResNet50}$. This asymmetry may arise because certain models can capture features learned by others, while the reverse is not true.
\begin{figure*}[tbh]
\centering
\begin{subfigure}[CLIP$_\text{ViT-B/32}$]
{\includegraphics[angle=0, width=0.23\textwidth]{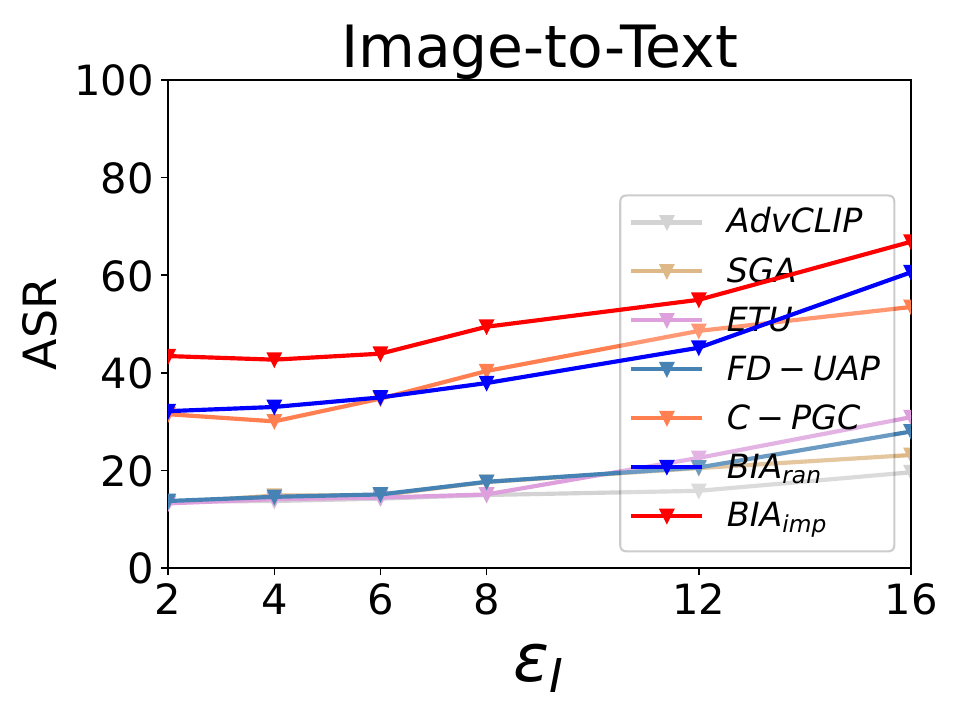}}
\end{subfigure}
\begin{subfigure}[CLIP$_\text{ViT-B/32}$]
{\includegraphics[angle=0, width=0.23\textwidth]{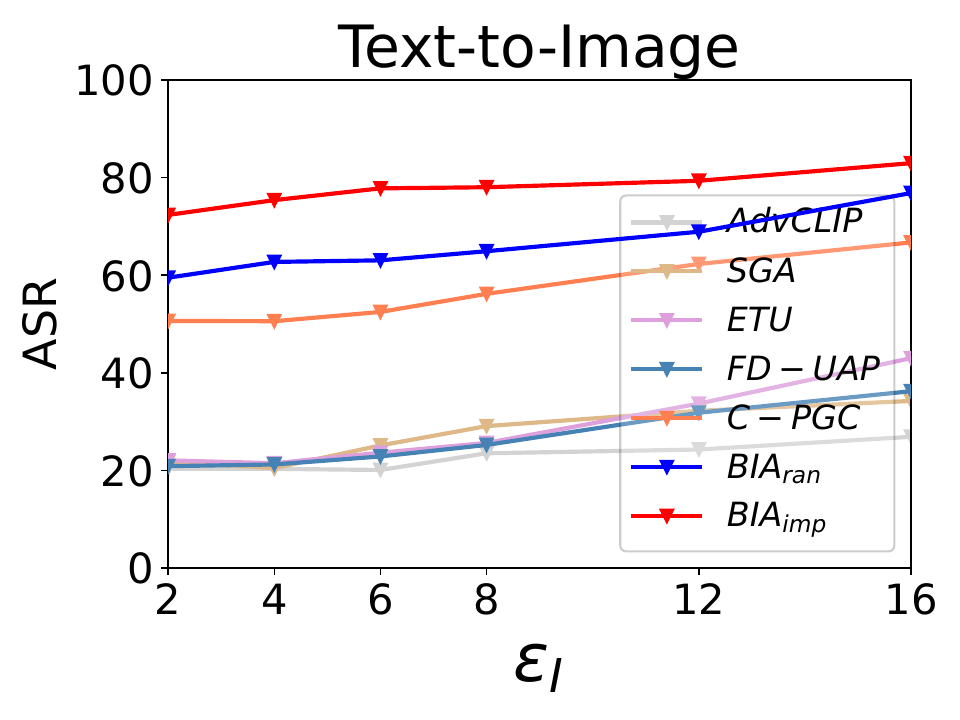}}
\end{subfigure}
\begin{subfigure}[RN50]
{\includegraphics[angle=0, width=0.23\textwidth]{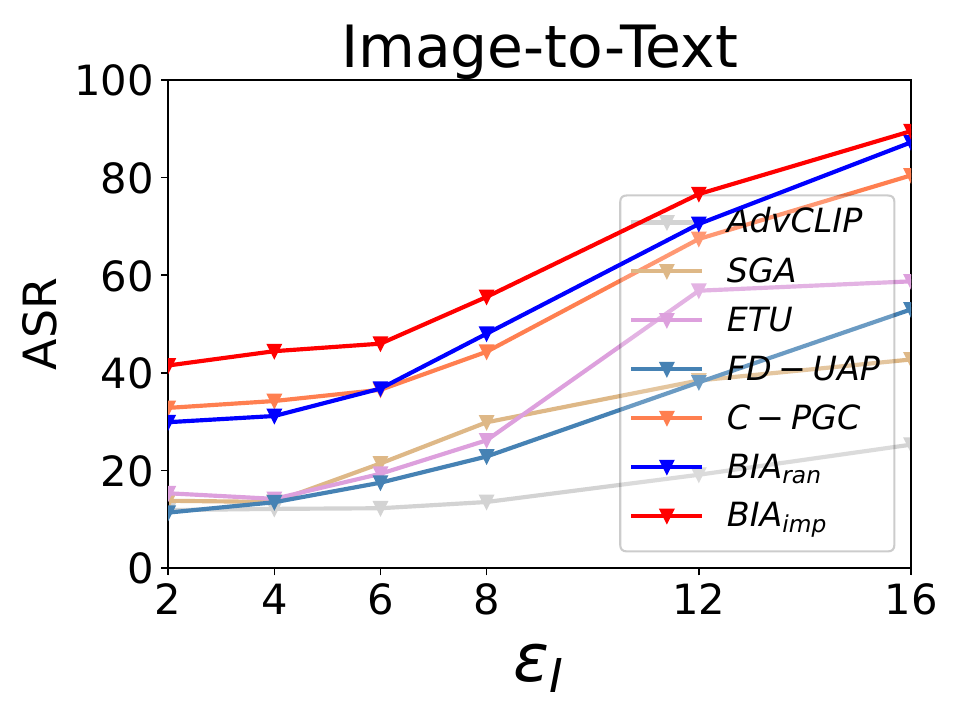}}
\end{subfigure}
\begin{subfigure}[RN50]
{\includegraphics[angle=0, width=0.23\textwidth]{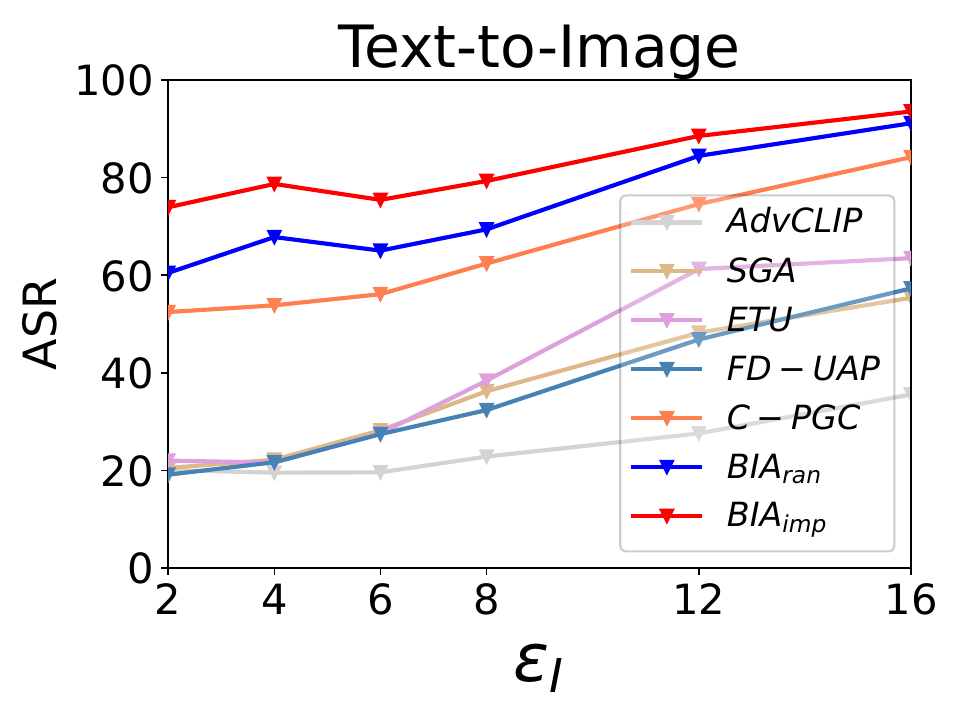}}
\end{subfigure}\\
\caption{The attack success rate $(\%)$ of R@1 in image-text retrieval on Flickr30K under different magnitudes of image UAPs. The source model is CLIP$_\text{ViT-B/16}$.}
\label{datamag_img}
\end{figure*}

\begin{figure*}[tbh]
\centering
\begin{subfigure}[CLIP$_\text{ViT-B/32}$]
{\includegraphics[angle=0, width=0.23\textwidth]{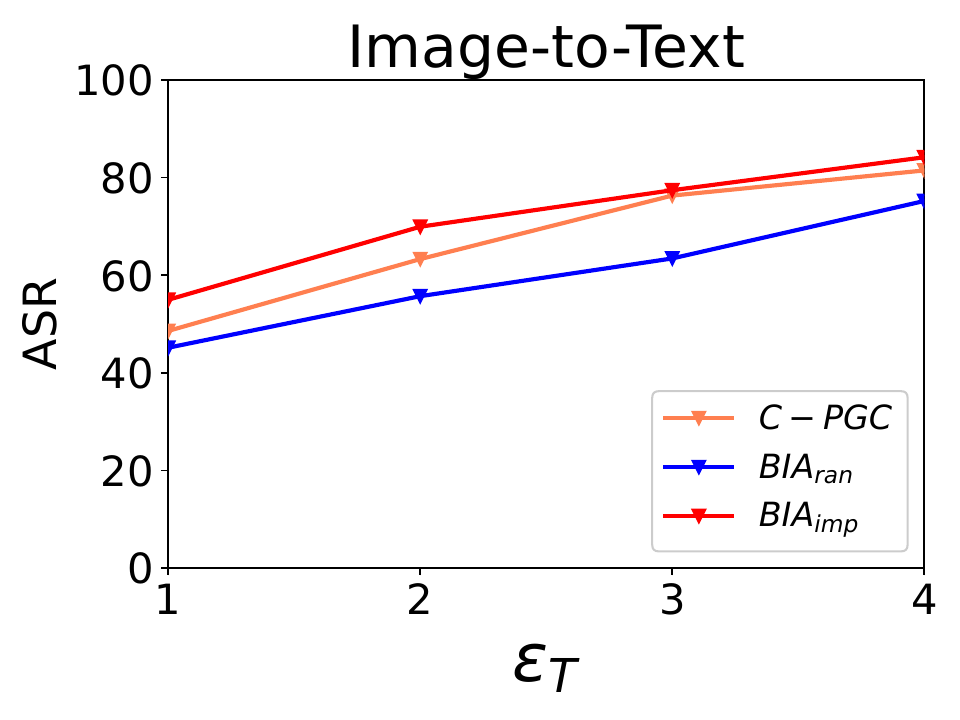}}
\end{subfigure}
\begin{subfigure}[CLIP$_\text{ViT-B/32}$]
{\includegraphics[angle=0, width=0.23\textwidth]{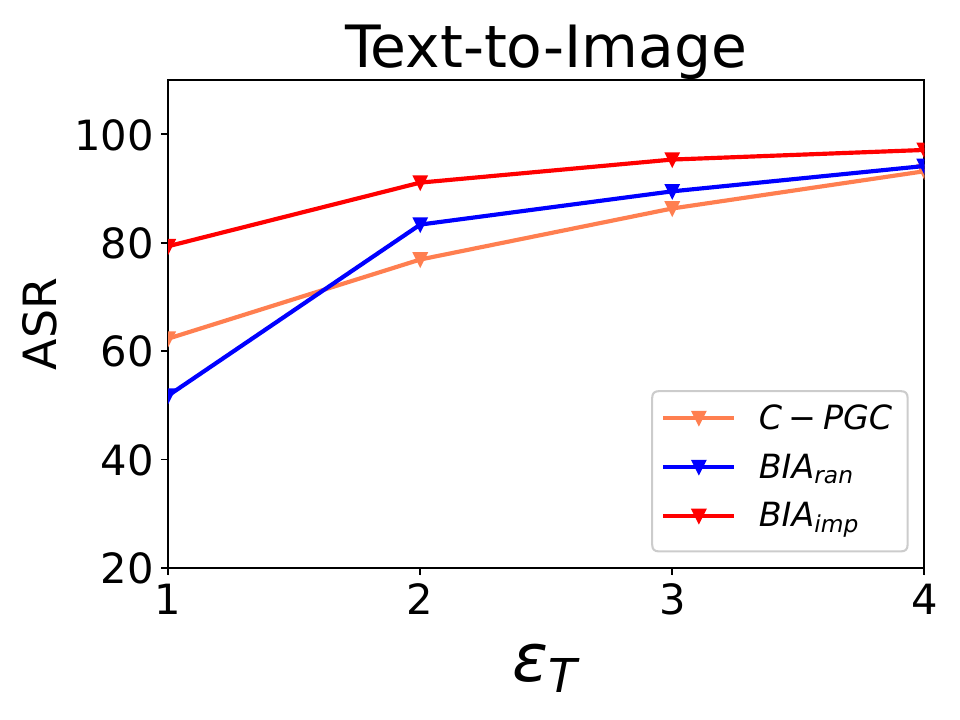}}
\end{subfigure}
\begin{subfigure}[CLIP$_\text{RN50}$]
{\includegraphics[angle=0, width=0.23\textwidth]{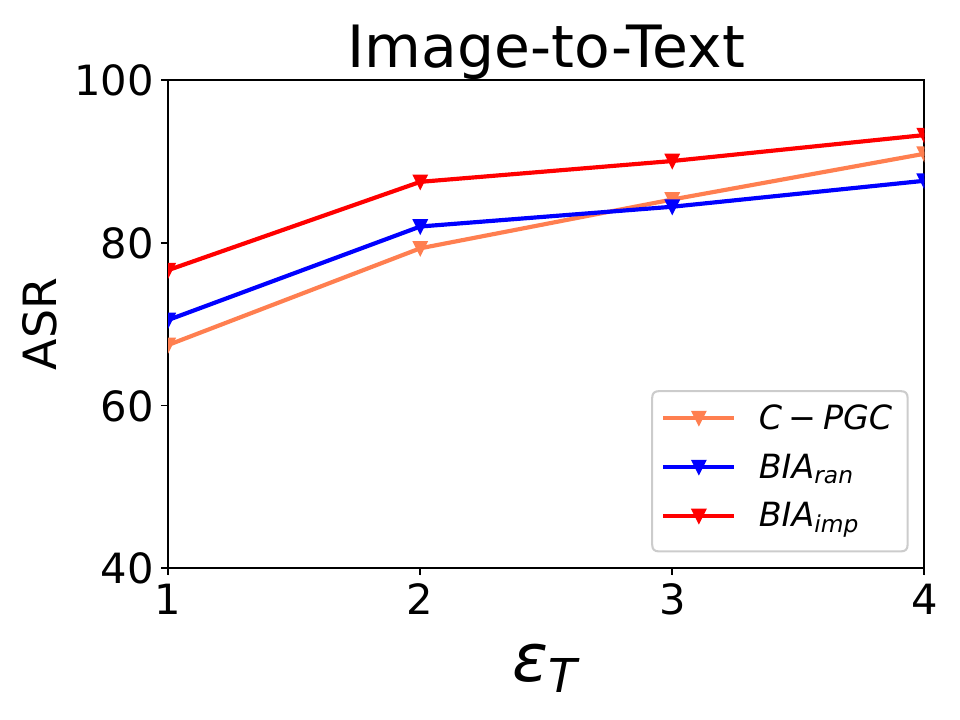}}
\end{subfigure}
\begin{subfigure}[CLIP$_\text{RN50}$]
{\includegraphics[angle=0, width=0.23\textwidth]{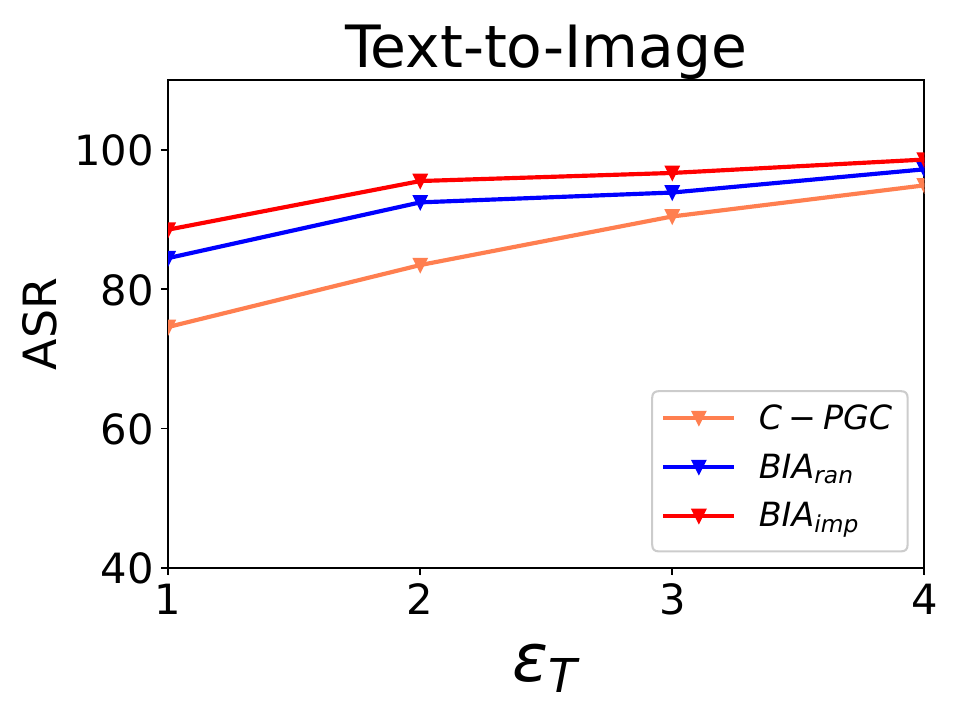}}
\end{subfigure}\\
\caption{The attack success rate $(\%)$ of R@1 in image-text retrieval on Flickr30K under different magnitudes of text UAPs. The source model is CLIP$_\text{ViT-B/16}$.}
\label{datamag_txt}
\end{figure*}

\subsection{Cross-task Transferability}

\begin{figure}[tbh]
\centering
\begin{subfigure}[CLIP$_\text{ViT-B/32}$]
{\includegraphics[angle=0, width=0.23\textwidth]{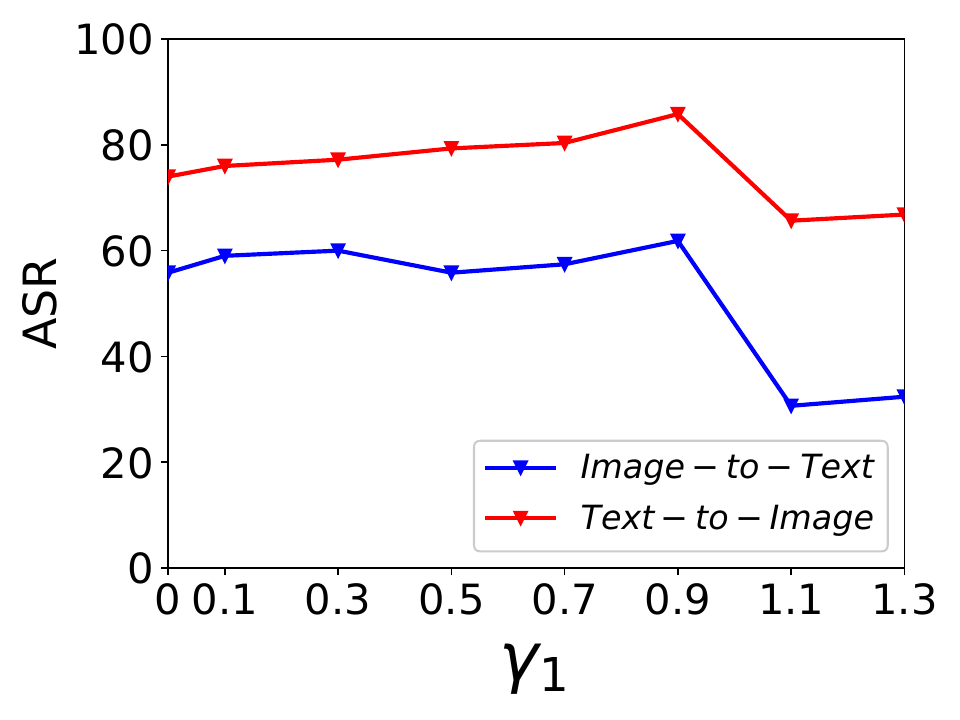}}
\end{subfigure}
\begin{subfigure}[CLIP$_\text{ViT-B/32}$]
{\includegraphics[angle=0, width=0.23\textwidth]{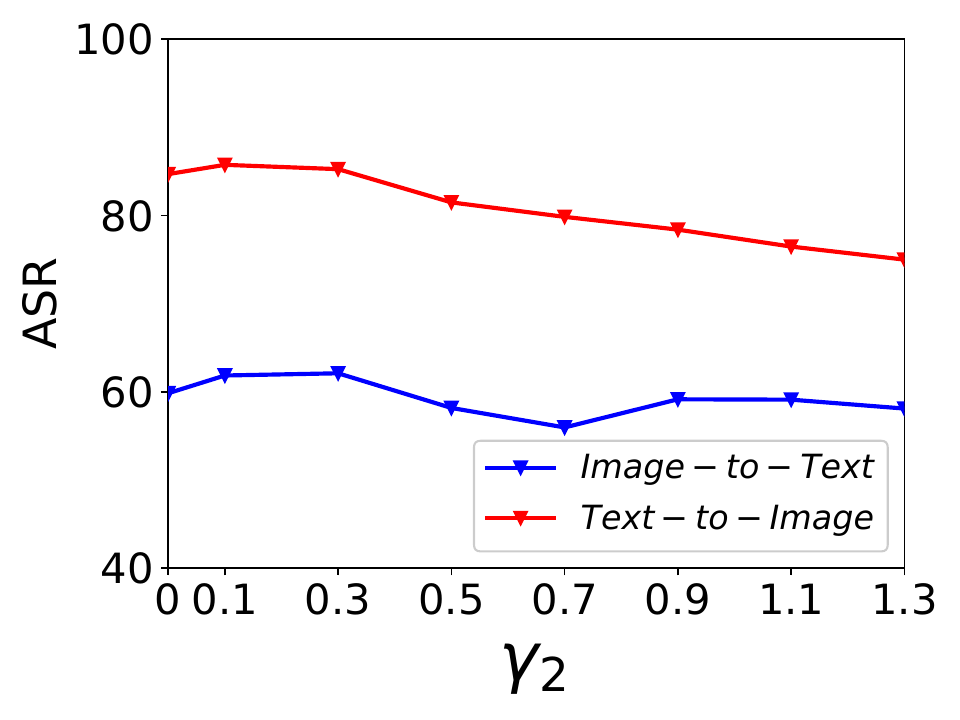}}
\end{subfigure}
\caption{The attack success rate $(\%)$ of R@1 in image-text retrieval on Flickr30K under different weights of past and future gradients. The source model is CLIP$_\text{ViT-B/16}$.}
\label{data_grad}
\end{figure}

We evaluate the effectiveness of the proposed method across different tasks, including image–text retrieval, visual grounding, and image captioning.

\begin{figure}[tbh]
\centering
\begin{subfigure}[CLIP$_\text{ViT-B/32}$]
{\includegraphics[angle=0, width=0.23\textwidth]{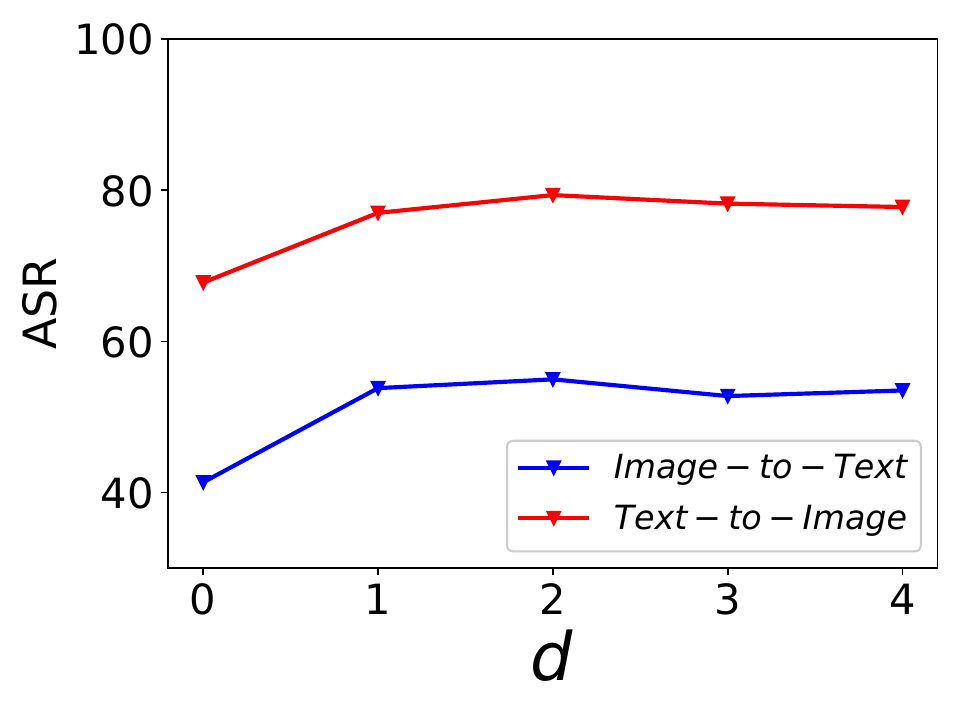}}
\end{subfigure}
\begin{subfigure}[CLIP$_\text{RN50}$]
{\includegraphics[angle=0, width=0.23\textwidth]{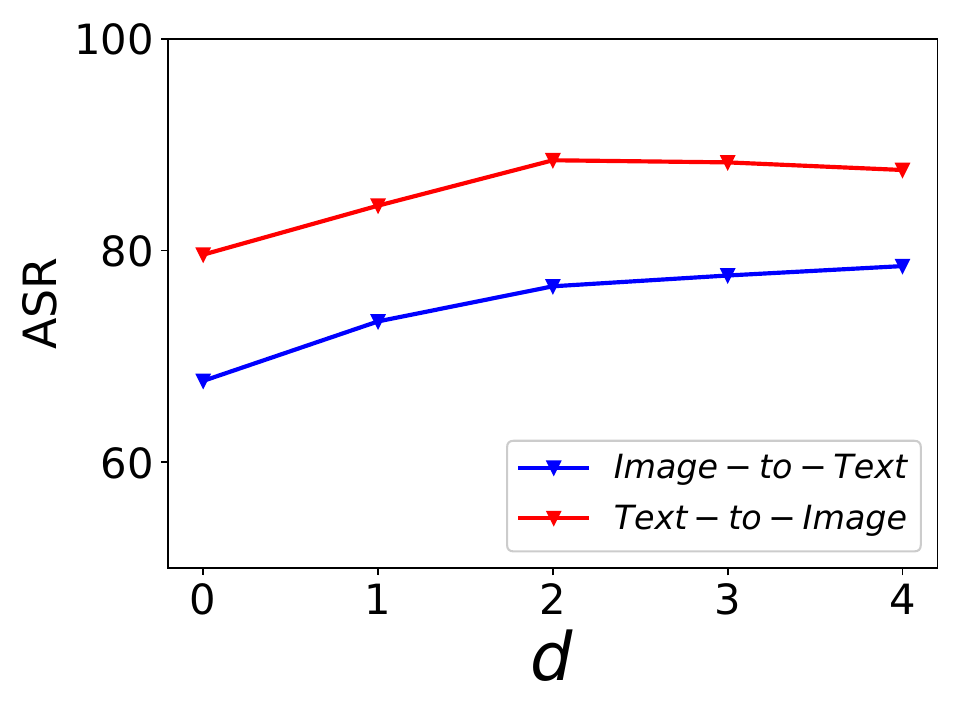}}
\end{subfigure}
\caption{The attack success rate in terms of the average of R@1 in image-text retrieval on Flickr30K under different numbers of future gradient steps. The source model is ViT-B/16-based CLIP.}
\label{futr_grad}
\end{figure}

\subsubsection{From image-text retrieval to other tasks} 
We first examine transfer attacks from image–text retrieval to visual grounding and image captioning, with results presented in Tables \ref{tab_trans_3} and \ref{tab_trans_4}, respectively. For both experiments, we generate UAPs using CLIP$_\text{ViT-B/16}$ and ALBEF as the source models and Flickr30K as the source dataset. For visual grounding, we use ALBEF and RefCOCO+ as the target model and dataset. For image captioning, we use BLIP and MSCOCO as the target model and dataset. 

Across both tasks, all methods exhibit notably reduced transferability. This is expected due to substantial differences in data distributions, model architectures, and task objectives. Different tasks encourage models to learn distinct feature representations and thus possess different vulnerabilities. For instance, image–text retrieval focuses on global cross-modal similarity, whereas image captioning requires fine-grained visual understanding and linguistic generation. Consequently, UAPs that degrade retrieval features may not effectively disrupt captioning models. Nevertheless, the proposed method achieves the strongest universal adversarial performance. This is attributed to its capability to prevent overfitting by considering past and future update directions and the incorporation of text attacks via hierarchical modeling of textual importance.

\begin{table*}[!t]
\center
\caption{Ablation study of different components on Flickr30K. The attack success rate of R@1 on
image-text retrieval is reported. CLIP$_\text{ViT-B/16}$ is adopted as the source model.}
\label{tab_abla}
\centering
\scalebox{0.9}{
\begin{tabular}{c|cc|cc|cc|cc|cc|cc|cc}
\toprule

\multicolumn{1}{c}{\multirow{2}{*}{\textbf{Target Model}}} & \multicolumn{10}{c}{\textbf{CLIP}} & \multicolumn{2}{c}{\multirow{2}{*}{\textbf{ALBEF}}} & \multicolumn{2}{c}{\multirow{2}{*}{\textbf{TCL}}}\\
\cmidrule(lr){2-11} 
&\multicolumn{2}{c|}{\text{RN50}} &\multicolumn{2}{c|}{\text{RN101}} & \multicolumn{2}{c|}{$\text{ViT-B/16}$}  & \multicolumn{2}{c|}{$\text{ViT-B/32}$}  & \multicolumn{2}{c|}{$\text{ViT-L/14}$} \\

\midrule
{ Method} & I2T & T2I & I2T & T2I  & I2T & T2I & I2T & T2I & I2T & T2I & I2T & T2I & I2T & T2I  \\
\midrule
{HRA$_{imp}$ w $\text{MixUp}$} &71.73  &86.64  &74.97 &88.22  &\cellcolor[HTML]{EFEFEF}{90.67} &\cellcolor[HTML]{EFEFEF}{96.81}  &54.85  &79.25   &61.29   &83.79 &40.46  &69.43     &51.90   &75.36 \\
{HRA$_{imp}$ w $\text{BSA}$}   &66.75  &80.88 &69.99 &85.21  &\cellcolor[HTML]{EFEFEF}{94.60} &\cellcolor[HTML]{EFEFEF}{94.46}   &50.34  &77.73  &55.71   &75.10  & 40.25  &68.92    &51.32 & 73.88 \\
{HRA$_{imp}$ w $\text{TIA}$}  &68.50  &82.32 &69.09  &85.76  &\cellcolor[HTML]{EFEFEF}{94.11} &\cellcolor[HTML]{EFEFEF}{88.72}    &51.46 &76.25  &54.36   &74.74  &38.27  &68.43     &52.05  &73.86 \\
{HRA$_{imp}$ w $\text{SIA}$}   &64.09  &82.39 &68.84  &85.63  &\cellcolor[HTML]{EFEFEF}{74.72} &\cellcolor[HTML]{EFEFEF}{95.04}    &52.02  &77.44     &49.57  &75.32  &37.64  &68.36 &50.47   &75.52 \\
{HRA$_{imp}$ w $\text{Mom}$} &71.86 &87.06   &70.85  &82.27  &\cellcolor[HTML]{EFEFEF}{35.71}  &\cellcolor[HTML]{EFEFEF}{69.68}     &52.58  &77.48 &59.83  &85.48  &36.13   &68.44     &49.30  &74.93 \\
{HRA$_{imp}$ w/o $\text{FM}$} &67.69 &79.62   &66.67  &80.51  &\cellcolor[HTML]{EFEFEF}{91.41}  &\cellcolor[HTML]{EFEFEF}{94.59}   &41.35  &67.78  &39.26   &60.82  &34.10   &68.50    &48.79  &75.98 \\
{HRA$_{imp}$ w/o $\text{Text attack}$}  &67.43  &76.35  &65.52   &74.52  & \cellcolor[HTML]{EFEFEF}{80.49} &\cellcolor[HTML]{EFEFEF}{89.34}  &28.34  &45.72  &36.81   &66.33   &20.33     &21.23   &24.34 &24.38   \\
{HRA$_{imp}$ w/o $\text{Image attack}$}  &40.49  &74.37 &41.63   &74.65 & \cellcolor[HTML]{EFEFEF}{36.32} &\cellcolor[HTML]{EFEFEF}{67.88}  &41.47  &64.84  &32.02   &61.56   &22.73     &57.95  &35.30 &69.12   \\
{HRA$_{imp}$}  &76.63  &88.54 &77.39  &88.39 & \cellcolor[HTML]{EFEFEF}{99.05} &\cellcolor[HTML]{EFEFEF}{95.27}  &54.97  &79.35  &61.84   &85.73   &40.98     &70.28  &52.37 &76.95  \\
\bottomrule

\end{tabular}
}
\end{table*}

\subsubsection{From visual grounding to image-text retrieval} 
We also evaluate transfer attacks from visual grounding to image–text retrieval, with results reported in Table \ref{tab_trans_5}. The consistent improvements across models further confirm the effectiveness of the proposed method.

In summary, the strong performance across cross-model and cross-task transfers on different datasets validates the effectiveness and transferability of our approach.

\subsection{Parameter analysis}
\subsubsection{Influence of Perturbation Budgets}
The perturbation magnitude plays a critical role in balancing attack effectiveness and imperceptibility. Figures~\ref{datamag_img} and~\ref{datamag_txt} present the results of image and text attacks under varying magnitudes. From the figures, it can be observed that attack performance generally improves as the magnitude of UAPs increases. However, excessively large perturbations can degrade visual quality and make attacks more detectable. Importantly, across all magnitudes, the proposed method consistently outperforms competing approaches, further highlighting its superiority.

\begin{figure*}[tbh]
\centering
\begin{subfigure}
{\includegraphics[angle=0, width=0.9\textwidth]{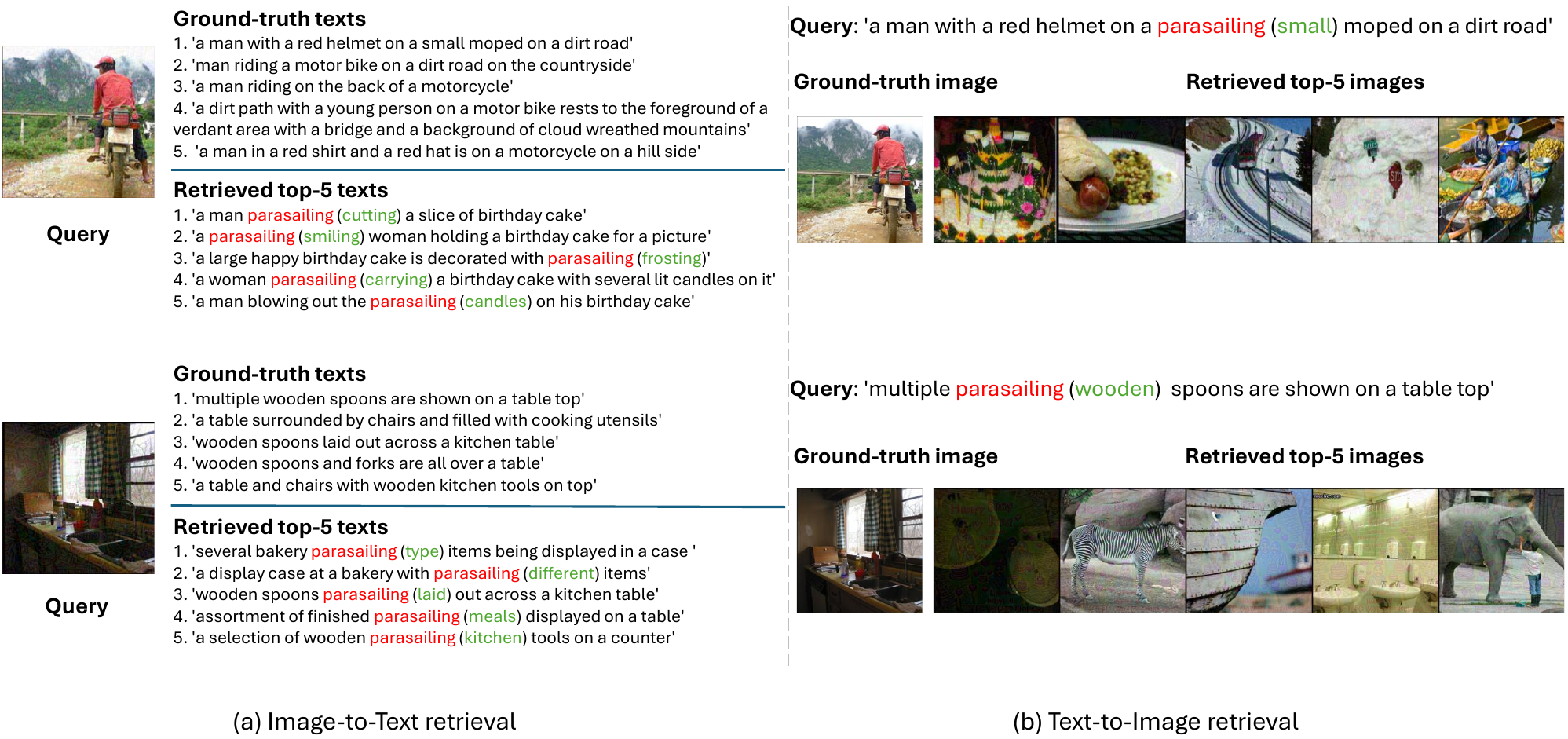}}
\end{subfigure}
\caption{Examples of top-5 image-text retrieval results. Red words indicate the text UAPs, while green words show the originals.}
\label{ITR}
\end{figure*}

\subsubsection{Influence of Combination Weights in Future-Aware Momentum}
We further analyze the contributions of past and future gradients by evaluating attack performance under different weighting schemes, using {HRA$_{imp}$} as an example. The corresponding results are shown in Figure~\ref{data_grad}, from which critical observations can be made. For past gradients, very small weights fail to sufficiently regularize the current update, whereas excessively large weights overly constrain the optimization to historical directions, reducing attack effectiveness. For future gradients, even small weights provide useful guidance by anticipating upcoming trends, while large weights can also lead to overfitting to the future optimization trend.

\subsubsection{Influence of the Number of Future Steps}
To investigate how future gradients contribute to improved transferability, we use {HRA$_{imp}$} as an example and report the attack performance in Figure~\ref{futr_grad}. The results show that a longer update horizon generally enhances performance. However, in some cases, an excessively long horizon begins to degrade it. For example, when attacking CLIP${\text{ViT-B/16}}$, extending the horizon from 1 to 2 steps improves transferability, whereas further increasing it from 2 to 4 steps leads to a slight performance drop. This degradation occurs because too many update steps cause the perturbation to overfit the source model’s gradients, thereby weakening its transferability to other models.

In terms of computational cost for incorporating future update information, on a workstation equipped with an NVIDIA L40s GPU, the training time of the proposed method for $0$, $1$, $2$, $3$, and $4$ steps is $229$, $309$, $553$, $675$, and $675$ seconds, respectively. Overall, using $2$ steps provides the best trade-off between performance and efficiency, with an acceptable computational overhead.

\begin{figure*}[tbh]
\centering
\begin{subfigure}[Image Attacks]
{\includegraphics[angle=0, width=0.48\textwidth]{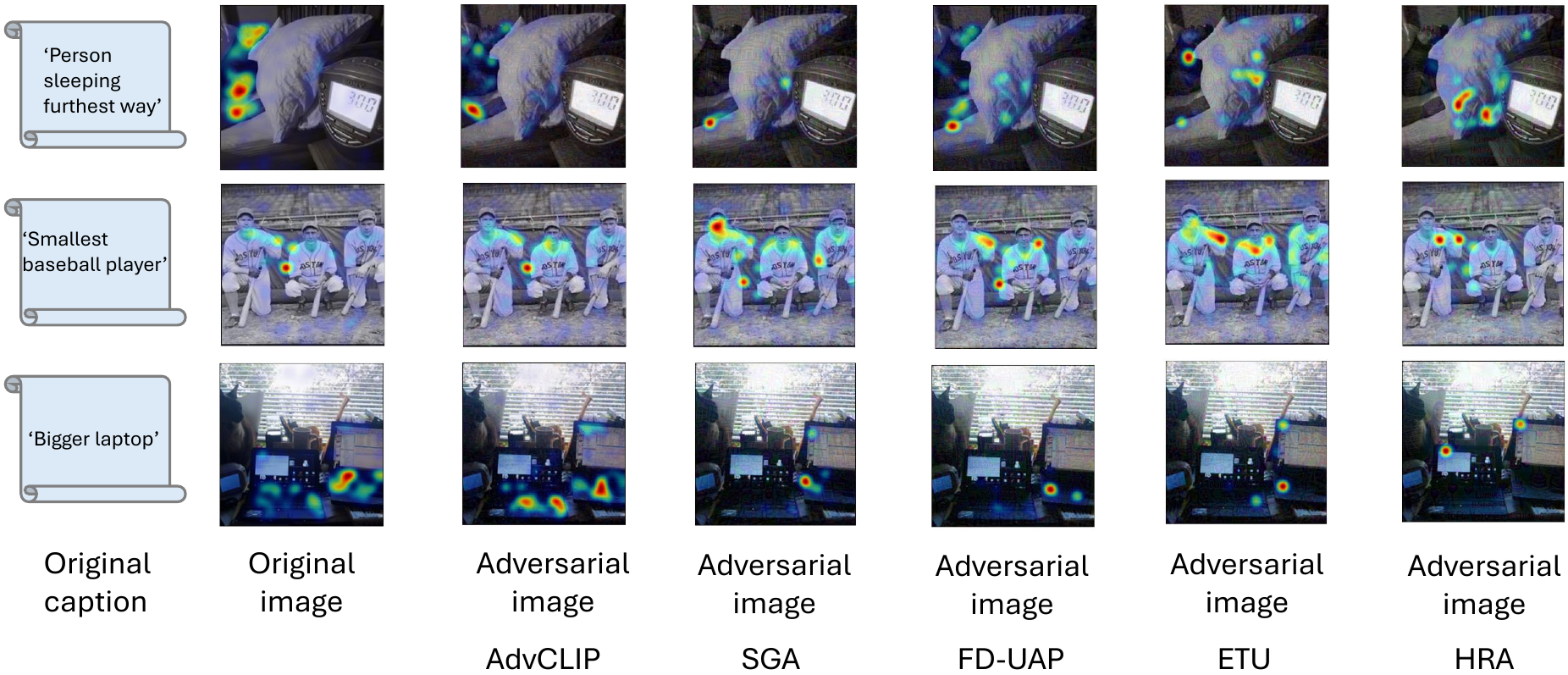}}
\end{subfigure}
\begin{subfigure}[Multimodal Attacks]
{\includegraphics[angle=0, width=0.48\textwidth]{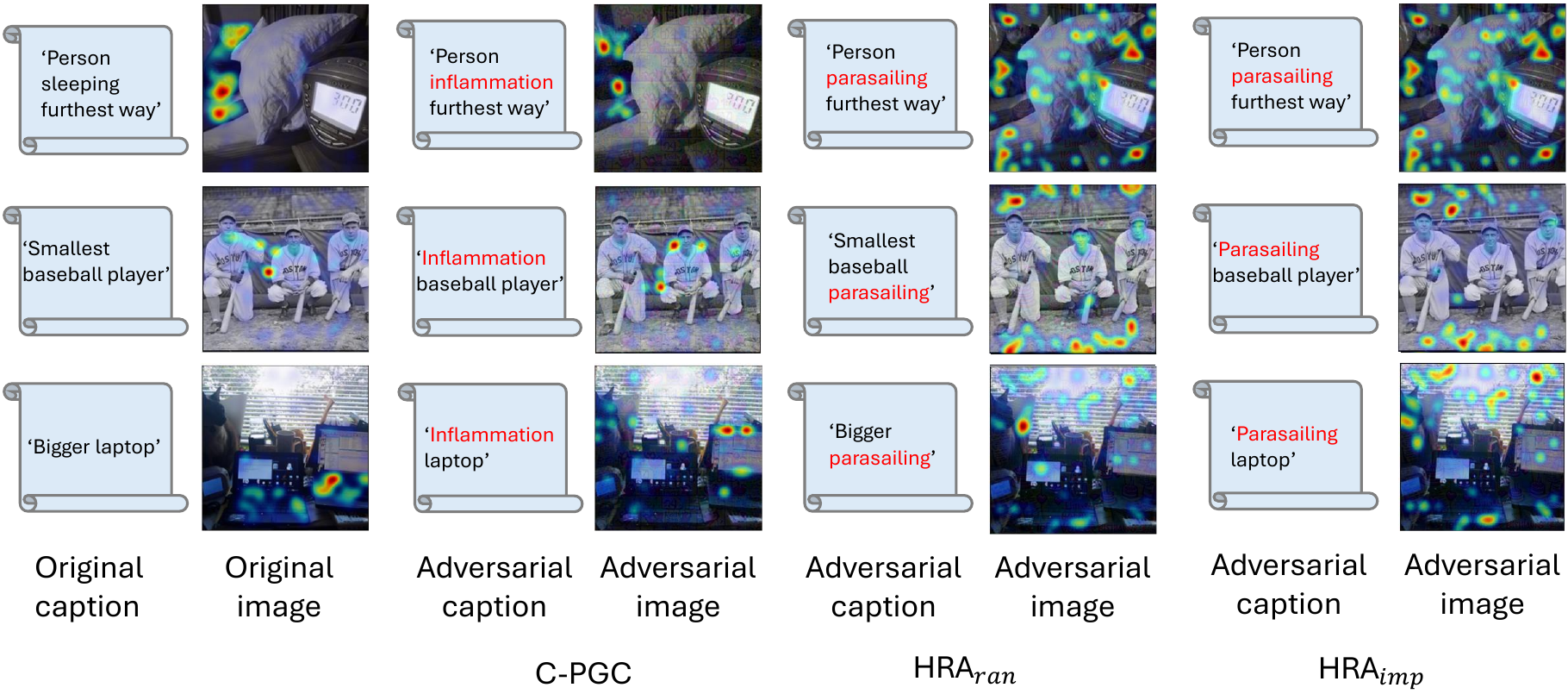}}
\end{subfigure}
\caption{The Grad-CAM visualizations of the original data and the perturbed data.}
\label{Grad}
\end{figure*}

\subsection{Ablation study}
To verify the effectiveness of each component of the proposed method, we conduct extensive ablation studies using {HRA${imp}$} as an example. First, we evaluate the contribution of individual components by removing them separately, resulting in {HRA${imp}$ w/o FM} (future-aware momentum), {HRA${imp}$ w/o Text attack}, and {HRA${imp}$ w/o Image attack}. Second, we assess our approach with several existing augmentation strategies, including Random Crop ({HRA w Crop}), MixUp ({HRA w MixUp}), Block-wise Augmentation \cite{wang2023structure} ({HRA w BWA}), Translation \cite{dong2019evading} ({HRA w TIA}), and Rescale \cite{lu2023set} ({HRA w SIA}). Third, we replace future-aware momentum with conventional momentum, denoted as {HRA w Mom}.

Table~\ref{tab_abla} summarizes the results of all variants under the image–text retrieval setting, using CLIP$_\text{ViT-B/16}$ and Flickr30K as the source model and dataset, respectively. The results indicate that each proposed component contributes positively to attack performance and consistently outperforms the corresponding alternatives. In addition, the proposed method can be combined with different augmentation strategies to further enhance performance. These findings further validate the effectiveness of the proposed method.

\subsection{Visualization}
Figure \ref{ITR} presents examples of image–text retrieval under the proposed attack. The results show that the method can effectively mislead the target model into returning incorrect retrievals while remaining visually imperceptible. dditionally, we present some Grad-CAM \cite{selvaraju2017grad} visualization examples in Figure \ref{Grad}, where it can be seen that the UAPs can significantly change the attention of the target model. These further confirm the effectiveness of the proposed approach.

\section{Conclusions, Limitations and Future Work}\label{CONCLUSION_future}
\subsection{Conclusions}\label{CONCLUSION}
In this paper, we propose a universal multimodal adversarial attack framework that systematically explores the vulnerabilities of VLP models and targets the specific characteristics of attacks in each modality. To improve adversarial transferability, we introduce a Hierarchical Refinement Attack (HRA) that hierarchically regularizes the optimization trajectory for image UAPs, mitigating overfitting caused by convergence to local optima. For the text modality, we design a practical and effective UAP learning strategy that accounts for both intra- and inter-sentence importance, ensuring globally influential word replacements. Extensive experiments across diverse VLP models, downstream tasks, and datasets demonstrate the effectiveness and transferability of the proposed attacks.

\subsection{Limitations and Future Work}\label{future}
Despite the significant improvements in universal adversarial transferability, the text attack that inserts a uniform word across different sentences remains perceptible to human readers. This limitation stems from the discrete nature of text, even when the perturbation magnitude is strictly constrained. In future work, we aim to develop more imperceptible text-based attack strategies. Moreover, adversarial transferability remains limited in certain scenarios, particularly under low perturbation budgets. We plan to further investigate the underlying mechanisms of model perception and explore more effective and transferable universal attack methods.

\clearpage
\bibliographystyle{IEEEtran}
\balance
\bibliography{acmart}

\end{document}